\documentclass[preprint,12pt]{elsarticle}

\usepackage[utf8]{inputenc} % allow utf-8 input
\usepackage[T1]{fontenc}    % use 8-bit T1 fonts
\usepackage{hyperref}
\usepackage{url}            % simple URL typesetting
\usepackage{booktabs}       % professional-quality tables
\usepackage{amsfonts}       % blackboard math symbols
\usepackage{nicefrac}       % compact symbols for 1/2, etc.
\usepackage{microtype}      % microtypography

\usepackage{amsmath,amssymb}
\usepackage{mathabx}
\usepackage{amsthm}
\usepackage{bm}
\usepackage{array}          % for \newcolumntype macro
\usepackage{mathptmx}       % Times Roman font for math
\usepackage{tablefootnote}
\usepackage{mathtools}
\usepackage[mathcal]{eucal}
\usepackage{graphicx}
\usepackage{subcaption}
\usepackage{color}

\usepackage{tabularx}

\usepackage{dsfont}

\newcommand{\todo}[1]{}
\renewcommand{\todo}[1]{{\color{red} TODO: {#1}}}

\usepackage{amssymb}
\renewcommand{\vec}[1]{\mathbf{#1}}
\renewcommand{\Re}{\mathbb{R}}

\DeclareMathOperator{\crps}{\textsc{CRPS}}
\DeclareMathOperator{\ncrps}{\textsc{N-CRPS}}

\DeclareMathOperator{\linear}{\textsc{Linear}}
\DeclareMathOperator{\fc}{\textsc{FC}}
\DeclareMathOperator{\relu}{\textsc{ReLu}}

\newcommand{\nbeatsinput}{\vec{x}}
\newcommand{\nbeatshidden}{\vec{h}}
\newcommand{\nbeatsbackcast}{{\widehat{\nbeatsinput}}}
\newcommand{\nbeatsforecast}{{\widehat{\vec{y}}}} 
\newcommand{\windowlength}{w}

\journal{Applied Energy}

\begin{document}

\begin{frontmatter}

%% Title, authors and addresses

%% use the tnoteref command within \title for footnotes;
%% use the tnotetext command for theassociated footnote;
%% use the fnref command within \author or \address for footnotes;
%% use the fntext command for theassociated footnote;
%% use the corref command within \author for corresponding author footnotes;
%% use the cortext command for theassociated footnote;
%% use the ead command for the email address,
%% and the form \ead[url] for the home page:
%% \title{Title\tnoteref{label1}}
%% \tnotetext[label1]{}
%% \author{Name\corref{cor1}\fnref{label2}}
%% \ead{email address}
%% \ead[url]{home page}
%% \fntext[label2]{}
%% \cortext[cor1]{}
%% \address{Address\fnref{label3}}
%% \fntext[label3]{}

\title{Any-Quantile Probabilistic Forecasting of Short-Term Electricity Demand}
% \title{Any-quantile approach for probabilistic short-term electricity demand forecasting}

%% use optional labels to link authors explicitly to addresses:
%% \author[label1,label2]{}
%% \address[label1]{}
%% \address[label2]{}

\author[a0]{Slawek Smyl}
\author[a1]{Boris N. Oreshkin\fnref{boris}}
\author[a2]{Pawe\l\  Pe\l ka}
\author[a3]{Grzegorz Dudek}

\address[a0]{Walmart, USA, e-mail: slaweks@hotmail.co.uk}
\address[a1]{Amazon, Canada, e-mail: boris.oreshkin@gmail.com}
\address[a3]{Department of Electrical Engineering,
Czestochowa University of Technology, 42-200 Czestochowa, Al. Armii Krajowej 17, Poland, e-mail: grzegorz.dudek@pcz.pl}
\address[a2]{Department of Intelligent Computer Systems,
Czestochowa University of Technology, 42-200 Czestochowa, Al. Armii Krajowej 36, Poland, e-mail:  pawel.pelka@pcz.pl}

\fntext[boris]{This work does not relate to author's position at Amazon}

\begin{abstract}
Power systems operate under uncertainty originating from multiple factors that are impossible to account for deterministically. Distributional forecasting is used to control and mitigate risks associated with this uncertainty. Recent progress in deep learning has helped to significantly improve the accuracy of point forecasts, while accurate distributional forecasting still presents a significant challenge. In this paper, we propose a novel general approach for distributional forecasting capable of predicting arbitrary quantiles. We show that our general approach can be seamlessly applied to two distinct neural architectures leading to the state-of-the-art distributional forecasting results in the context of short-term electricity demand forecasting task. We empirically validate our method on 35 hourly electricity demand time-series for European countries. 
% Our code will be made publicly available should the paper be accepted.
Our code is available here: \url{https://github.com/boreshkinai/any-quantile}.

\end{abstract}

% %%Graphical abstract
% \begin{graphicalabstract}
% \includegraphics[width=\textwidth]{fig/AQ_concept.pdf}
% \end{graphicalabstract}

% %%Research highlights
% \begin{highlights}
% \item New probabilistic forecasting approach can predict any quantile 
% \item Any-quantile ESRNN architecture is proposed
% \item Any-quantile N-BEATS architecture is proposed
% \item Any-quantile approach outperforms statistical and neural baselines
% \end{highlights}

\begin{keyword}
short-term load forecasting \sep probabilistic forecasting \sep quantile regression \sep deep learning
%% keywords here, in the form: keyword \sep keyword

%% PACS codes here, in the form: \PACS code \sep code

%% MSC codes here, in the form: \MSC code \sep code
%% or \MSC[2008] code \sep code (2000 is the default)

\end{keyword}

\end{frontmatter}

%% \linenumbers

%% main text
\section{Introduction}
\label{Intro}

Extraneous stochastic factors, which include weather fluctuations, load variations and equipment failures, make probabilistic forecasting arguably the only viable approach to forecasting in power systems~\cite{Hau19}. Concretely, these factors introduce uncontrollable random variability in power system operations, which can only be accounted for, at any particular point in time, by keeping a model of many possible futures using the tools of probabilistic calculus. Probabilistic calculus induces probabilistic forecasts that provide likely distribution of electricity demand for a specific point in the future or a set of likely demand trajectories in the context of multi-horizon forecasting. Power system operators can use these probabilistic forecasts to enhance operational efficiency, mitigate risks and ensure reliable and secure power supply. This correlates with the following range of the applications of forecasting in power systems: (i) planning and optimization, (ii) security assessment and management and (iii) energy trading. In the following, we discuss the motivations behind the use of probabilistic forecasting in these application areas in more detail.

\textbf{Planning and optimization} benefit from probabilistic forecasting by incorporating probabilistic forecasts of load demand and generation availability to minimize operating costs, maximize revenue and/or ensure reliable power supply via stochastic optimization and scheduling. For example, operating reserve can be planned for uncertainties to maintain grid reliability during peak demand periods, unexpected events or sudden changes in weather conditions. Concretely, the integration of probabilistic forecast information can reduce the predicted regulation range by about 12-31\% \cite{Eti18}, which reduces the necessary reserve margin costs and helps better utilize limited system balancing resources. Wang et al.~\cite{Wan23} show that the probabilistic forecast information can be transformed into dynamic reserve requirements, which is beneficial in terms of reducing the cost and the number of power balance and reserve violations compared to the static reserve requirements. Finally, probabilistic renewable energy generation forecasts enable the design of ramping products that provide flexible and efficient resources to balance the grid during periods of changing generation patterns, ensuring the seamless integration of renewable energy into the power system~\cite{Yam22}.

\textbf{Security assessment and management} in power systems focus on identifying high-risk extreme events and applying preventive measures accordingly~\cite{Le16}. For example, probabilistic forecasts of electricity demand can be used to identify high-likelihood future load demand increase events and operators can take preventive measures such as activating additional generation resources, adjusting generation schedules, or implementing load shedding plans to ensure grid stability. Other stability risks such as voltage fluctuations, frequency deviations and power imbalances, can also be better managed using probabilistic forecasts.

\textbf{Energy trading} optimizes generation and consumption patterns, manages risks and maximizes economic benefits by buying and selling electricity in various markets, such as wholesale electricity markets, balancing markets and ancillary services markets~\cite{Bey22}. Probabilistic forecasting provides valuable insights to energy traders for optimizing their trading strategies. It quantifies the range of possible outcomes and associated uncertainties in key parameters, such as electricity prices, load demand and generation availability, which are essential for trading decisions. For example, if a probabilistic load forecast indicates a higher probability of a peak load scenario, traders can adjust their generation schedules or contract additional capacity to meet the expected demand.

In our current work, we focus on the short-term electricity demand forecasting, which has prominent applications in energy trading as well as short-term planning and scheduling optimization. The rest of the section provides the review of the most relevant literature in probabilistic forecasting and its applications in power systems, as well as summarizes core motivation and contributions of our work. 

\subsection{Related Work} \label{ssec:related_work}

\textbf{Probabilistic forecasting in power systems} has profound impact on operational security, results in quantifiable financial outcomes and as such has profound impact on decision making in energy sector as a whole. The concept of probabilistic load forecasting is comprehensively examined in \cite{Hon16}, focusing on the progress in general methodology and the impact in practical applications. Practical applications encompass such areas as probabilistic load flow, unit commitment and reliability planning. In contrast to point forecasting, probabilistic load forecasting characterizes load variations by providing predictions in the form of probability density functions (PDFs), quantiles, or prediction intervals (PIs). Practical examples of the use of forecast uncertainty in grid management are illustrated in~\cite{Hau19}. These include the employment of probabilistic data for grid security calculations, the optimization of unit commitment, the mitigation of risks associated with extreme conditions and events. Furthermore, the review of probabilistic forecasting in smart grid applications can be found in~\cite{Kha22}, while applications of probabilistic forecasting in solar power systems are reviewed in~\cite{Li20}.

\textbf{Probabilistic forecasting approaches}, as delineated by Hong et al.~\cite{Hon16}, can be broadly subdivided into three primary categories: (i) input scenario simulation, (ii) post-processing involving residual simulation or forecast combination and (iii) model-dependent interval construction and probabilistic forecasting models. In the first category, multiple scenarios for input variables, such as temperature, are employed to generate various point forecasts. These forecasts are then aggregated to create a probabilistic forecast. For example, \cite{Hyn10} employs semi-parametric additive models to estimate the relationships between demand and various driver variables, including temperatures, calendar effects and some demographic and economic variables. Subsequently, the distributions of annual and weekly peak electricity demands are forecasted by utilizing a mixture of temperature simulations, anticipated future economic scenarios and residual bootstrapping. In another approach, Taylor and Buizza~\cite{Tay02} use neural networks (NNs) for estimating probabilistic load forecasts, with weather ensemble predictions for multiple scenarios being integrated as input variables. Additionally, \cite{Kho19} introduces generic scenario-based probabilistic load forecasting models using an ensemble of regression trees. In this case, quasi-periodic time-series data are represented as matrices and the singular value decomposition is employed to generate temperature scenarios. Finally, \cite{Xie18} explores and empirically compares three basic methods for generating temperature scenarios: the fixed-date, shifted-date and bootstrap.

The second approach to generating probabilistic forecasts involves post-processing existing point forecasts. This is typically accomplished by applying the probability density function of residuals on top of existing point forecasts. For example, a common approach consists in fitting a normal distribution to forecasting errors~\cite{Xie17}. The key takeaway of the study is that the quality of the underlying model plays a crucial role in ensuring the success of post-processing probabilistic load forecast. The literature also includes more advanced methods of residual simulation, such as the application of GARCH~\cite{Hor06}. Moreover, residual simulation can be successfully combined with predicted model uncertainties as shown in~\cite{Cao20}. This requires a more sophisticated modeling approach employing a hybrid ensemble of deep belief networks, bagging, boosting approaches and a $k$-nearest neighbor classifier. Similarly, ensemble forecasting can be used as a standalone approach to generating probabilistic forecasts: a set of point forecasts produced by individual models can be transformed into a PI via a quantile regression as shown in~\cite{Liu17}.

The third probabilistic forecasting category encompasses forecasting models capable of generating probabilistic forecasts directly. Bayesian non-parametric techniques such as quantile regression based on Gaussian process are popular in this category. For example, in~\cite{Yan18} Gaussian process predicts load quantiles in the next time step and the probability density function is obtained through kernel density estimation (KDE). Similar forecasting approaches, involving feature selection, quantile modeling and KDE are employed in~\cite{He19,Liu22,He22}. The lasso method acts as feature selector, while neural network models quantiles via quantile regression in~\cite{He19}. \cite{Liu22} utilizes a copula model for feature selection, while combining a 1D Convolutional Neural Network (CNN) and a Gated Recurrent Unit (GRU) to produce quantile forecasts. In~\cite{He22}, quantile forecasts produced by different quantile regression neural models are transformed into PDFs through KDE. These PDFs are subsequently combined using a multilayer Gaussian mixture distribution. Li et al.~\cite{Li23} extend~\cite{Liu22} by additionally incorporating a time attention mechanism. This model additionally allows for quantitative modeling of two types of uncertainty: epistemic and aleatoric, which is achieved through the use of Monte Carlo dropout and an improved loss function based on the Kullback–Leibler divergence. Similar KDE-based approaches include \cite{He17,He22a,Dud24}. He et al.~\cite{He17} propose a kernel-based support vector quantile regression model using Copula theory, which can quantify the uncertainty between inputs, such as power load and real electricity price. The forecasting results under different quantiles are fed into the KDE function. KDE can also be used on top of basic learners as a part of the meta-learning approach~\cite{He22a}. Here, a quantile regression LSTM meta-learner combines point forecasts produced by tree-based models, including random forest, gradient boosting decision tree and light gradient boosting machine. The PDF is determined using KDE modified by Gaussian approximation of quantiles. Similarly, a quantile meta-model, in the form of quantile linear regression or quantile random forest, can be constructed by stacking on top of the array of base models as shown in~\cite{Dud24}.

\textbf{Parametric and non-parametric probabilistic forecasting.} In addition to the mechanism used to construct the forecast, probabilistic forecasting approaches can be categorized into parametric, semi-parametric and non-parametric~\cite{Hon20}. All of the models discussed thus far are representatives of the non-parametric family. Further examples of non-parametric methods are provided below. Their common trait is the use of advanced machine learning models in combination with the variants of pinball loss for supervision. For example,~\cite{Wan19} employs a Long Short-Term Memory (LSTM) network,~\cite{Zha19} uses feed-forward network and~\cite{Li23a} combines GRU and temporal fusion transformer, while all of them are supervised with pinball loss. Additionally,~\cite{Li23a} adds quantile constraints and PI penalty terms to to prevent quantile crossover and aid in constructing tighter PIs. Smyl et al.~\cite{Smy23} combine exponential smoothing with a recurrent neural network (RNN) based on a novel recurrent cell, specifically designed to efficiently model both short and long-term dependencies within time-series data. The resulting probabilistic forecasts are presented in the form of three quantiles, corresponding to the median and the bounds of PI. Alfieri and De Falco~\cite{Alf20} decompose the load time-series through wavelet decomposition and predict its components using quantile regression forest. The final load forecasts are reconstructed by interpolating quantile predictions to prevent quantile crossing. 

A parametric approach to generating probabilistic forecasts assumes a certain PDF for the forecast distribution and involves a forecasting model that produces parameters of this PDF. In its simplest form, PDF can be represented by a normal distribution with mean and variance parameters. Gaussian processes are a viable option for slightly more general modeling~\cite{Mee18}. A more heavy-tail model such as Student-t distribution can be used on top of MLP, DeepAR and Transformer models, as done in GluonTS library~\cite{Ale20}. 
Another approach to handling multi-modality and asymmetry of forecast distribution is based on Gaussian mixture model~\cite{Bru22}.

\subsection{Motivation and Contributions}

It is crucial to acknowledge the growing array of factors contributing to significant uncertainty in the energy sector. These encompass the dynamic expansion of distributed renewable energy sources, the evolution of smart grids, the advancement of energy storage systems, the increasing prevalence of electric vehicles, the adoption of extended demand response programs and the liberalized markets with increasingly intricate pricing policies. These factors must contend with sudden weather changes and extreme weather events, further complicating the landscape. The value of probabilistic forecasting in the context of tackling these challenges in power systems applications has been clearly demonstrated in the literature. At the same time, there is a clear gap in the probabilistic forecasting accuracy, which is primarily related to (i) the prevalent use of simple non-state-of-the-art architectural backbones (LSTM/GRU/CNN) that lag behind in terms of point forecasting accuracy and (ii) the prevalent use of simple quantile regression training and inference mechanisms that lag behind in terms of probabilistic forecasting accuracy. In this paper we close this gap by showing that a combination of state-of-the-art neural architectures with the advanced any-quantile distributional training approach leads to noticeable gain in point and probabilistic forecasting accuracies. In a nutshell, our research contributions can be summarized as follows:
\begin{enumerate}
    \item We introduce a novel approach to probabilistic forecasting. The approach is applicable to a variety of machine learning models. It enables a model to produce an arbitrary quantile at inference time.
    \item We propose architectural modifications to the ESRNN forecasting algorithm expanding it with the any-quantile capability.
    \item We propose architectural modifications to the N-BEATS forecasting algorithm expanding it with the any-quantile capability.
    \item We present empirical results for short-term electricity demand forecasting application, showing the superior accuracy of any-quantile enabled ESRNN and N-BEATS forecasters with respect to baselines, including both statistical and state-of-the-art neural models.    
\end{enumerate}

The rest of the paper is organized as follows. Section~\ref{sec:framework} formulates the problem and conveys analysis results. Section~\ref{Method} describes proposed neural architectures solving any-quantile forecasting task. Section~\ref{sec:results} describes our empirical findings. Finally, Sections~\ref{Discussion} and~\ref{Con} discuss our findings and conclude the paper.

\section{Any-Quantile Forecasting Framework} \label{sec:framework}

\begin{figure}[t]
    \centering
    \includegraphics[width=0.8\linewidth]{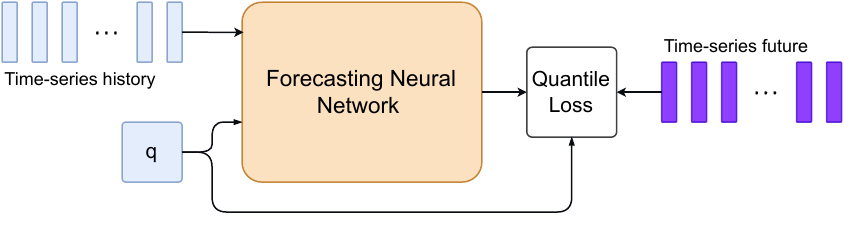}
    \caption{Any-quantile forecaster training methodology. The network accepts time-series history and a quantile level. The quantile level is generated randomly for each training sample, according to a pre-specified distribution (\emph{e.g.} uniform or Beta) and is used both as neural network input and as the supervision signal in the quantile loss.}
    \label{fig:any_quantile_overall}
\end{figure}

In this section we first describe the general statement of the short-term probabilistic forecasting problem. Then we outline the proposed general solution to this problem based on training a machine learning model using any-quantile approach as shown in Figure~\ref{fig:any_quantile_overall}. We further provide the analysis outlining the relationship between the mini-batch any-quantile loss and the CRPS metric.

\subsection{Short-term Distributional Forecasting Problem} 

We define the forecasting problem with a forecast horizon of length $H$ and a historical time-series $[y_1, \ldots, y_T] \in \Re^T$. Forecaster function $f_{\theta}: \Re^{\windowlength} \rightarrow \Re^{H}$, parameterized with $\theta \in \Theta$, predicts the vector of $q$-th quantiles of future values $\vec{y} \in \Re^H = [y_{T+1}, \ldots, y_{T+H}]$ based on past observations. For simplicity, we consider a \emph{lookback window}, $\nbeatsinput \in \Re^w = [y_{T-w+1}, \ldots, y_T]$ of length $w \le T$, which ends with the last observed value $y_T$. The indices of this window belong to the index set $\Delta^{in}_T = \{T-w+1, \ldots, T\}$. The forecast of the $q$-th quantiles of $\vec{y}$ is denoted as $\widehat{\vec{y}}_q$. The accuracy of distributional forecasting is assessed using Continuous Ranked Probability Score defined as:
\begin{equation} \label{eqn:crps_l2}
\crps(\widehat{F},y_o) = \int_\Re
       \left(\widehat{F}\left(y \right)- \mathds{1}{\{y \geq y_o\}} \right)^{2}\textrm{d}y \,,
\end{equation}
where $y_o$ is the observed value, $\widehat{F}$ denotes the empirical predictive cumulative distribution function (CDF) determined from the predicted set of quantiles and 
$\mathds{1}$ denotes the indicator function.

\subsection{Any-Quantile Learning}

The proposed general methodology enabling any-quantile neural network forecasting is inspired by \cite {gouttes2021probabilistic} and is depicted in Figure~\ref{fig:any_quantile_overall}.
The key feature of any-quantile forecasting is that both the neural network and the loss function accept quantile level $q$ as input. Suppose $y$ is observed value, $\widehat{y}_q$ is the predicted $q$-th quantile and the neural network is trained using quantile loss, defined as:
\begin{equation}
\rho(y, \widehat{y}_q) =
\begin{cases}
(y-\widehat{y}_q)q       & \text{if } y \geq \widehat{y}_q\\
(y-\widehat{y}_q)(q-1)  &\text{if } y < \widehat{y}_q 
\end{cases}.
\label{eqn:quantile_loss}
\end{equation}
Then the learning procedure defined based on this framework will possess interesting theoretical properties. In the following analysis we consider for simplicity and without loss of generality that $\nbeatsinput$ is the observed time-series history and $y$ is the unobserved scalar future value that we would like to forecast using a neural forecaster $f_{\theta}$ trained on a $N$-sample dataset of $(\nbeatsinput, y)$ tuples derived from the unknown joint probability density function $p(\nbeatsinput, y)$. Assuming that at training time quantile probability $q$ is generated uniformly at random in $[0,1]$ in the mini-batch of size $B$ and the following stochastic gradient descent (SGD) update, based on the quantile loss, is used to update neural forecaster parameters $\theta$ at iteration $k$:
\begin{equation}
\theta_{k+1} = \theta_{k} - \eta \nabla_{\theta} \frac{1}{B}\sum_i^B \rho(y_i, f_{\theta}(\nbeatsinput_i, q_i)),
\label{eqn:mini_batch_update}
\end{equation}
then the sequence of $\theta$ updates linearly converges to the solution of optimization problem on the full training dataset of size $N$~\cite{Karimi2016linear}:
\begin{equation}
\theta^{*} = \arg\min_{\theta \in \Theta} \frac{1}{N} \sum_i \rho(y_i, f_{\theta}(\nbeatsinput_i, q_i))\,.
\label{eqn:theta_optimal}
\end{equation}
By the strong law of large numbers, as $N$ increases without bound, the sum in the last equation converges to the following with probability 1:
\begin{equation}
\mathbb{E}_{\nbeatsinput, y} \mathbb{E}_{q} \rho(y, f_{\theta}(\nbeatsinput, q)) = \mathbb{E}_{\nbeatsinput, y} \int_{0}^{1} \rho(y, f_{\theta}(\nbeatsinput, q)) dq\,.
\label{eqn:sum_to_integral}
\end{equation}
Recalling that CRPS can be alternatively defined via the inverse CDF $F^{-1}$ of $y$ via quantile loss integral of the form~\cite{tilmann2011comparing}:
\begin{equation}
\crps(\widehat{F}, y) = 2 \int_{0}^{1} \rho(y, F^{-1}(q)) dq\,,
\label{eqn:crps_expected}
\end{equation}
we can clearly see a link between the mini-batch update~\eqref{eqn:mini_batch_update}, corresponding expected optimization loss~\eqref{eqn:sum_to_integral} and the integral form definition of CRPS~\eqref{eqn:crps_expected}.

\section{Any-Quantile Forecasting Models} \label{Method}

This section describes our main architectural contributions, showing the application of the general any-quantile forecasting methodology outlined in Section~\ref{sec:framework} to two popular neural forecasting architectures: ESRNN~\cite{smyl2022dyn} and N-BEATS~\cite{Ore19}. 

\subsection{AQ-ESRNN}

The any-quantile enabled ESRNN architecture, which we present in this section and call AQ-ESRNN is derived from ESRNN~\cite{smyl2022dyn}. Hence, preprocessing,  postprocessing and the general architectural approach of AQ-ESRNN and ESRNN are very similar. Therefore, in our current exposition we provide only necessary details and refer interested reader to~\cite{smyl2022dyn} for a more in-depth elucidation. 

\begin{figure}[t]
    \centering
    \includegraphics[width=1\linewidth]{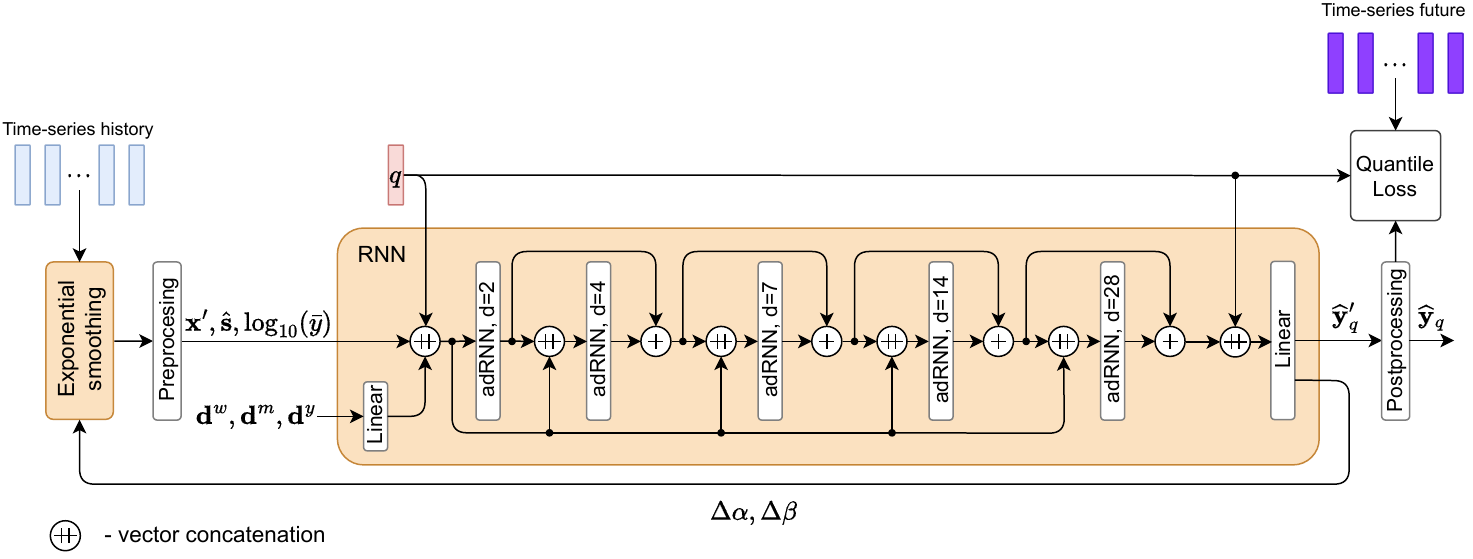}
    \caption{AQ-ESRNN architecture.}
    \label{fig:AQESRNN}
\end{figure}

\textbf{Exponential smoothing and pre-processing}. In the context of the short-term electricity forecasting, we observe that the hourly time-series exhibit several seasonalities including daily (24 hours), weekly (7 days, 168 hours), yearly (changing seasons). The high-frequency seasonalities are explicitly treated in AQ-ESRNN by the exponential smoothing and pre-processing blocks shown in Fig. \ref{fig:AQESRNN} using the $w=168$ point lag, while the yearly seasonality is dealt with by appending to the input an embedded week number. The exponential smoothing block is based on a version of Holt-Winters multiplicative Exponential Smoothing (ES) model~\cite{Hyn20}:
\begin{equation}
\begin{aligned}
l_{t,\tau}=\alpha_t \frac{y_\tau}{s_{t,\tau}} + (1-\alpha_t)l_{t,\tau-1}\,, \\
s_{t,\tau+w}=\beta_t \frac{y_\tau}{l_{t,\tau}} + (1-\beta_t)s_{t,\tau}\,,
\label{eqls1}
\end{aligned}
\end{equation} 
where $y_\tau$ is the system load at hour $\tau$, $l_{t,\tau}$ is a level component, $s_{t,\tau}$ is a weekly seasonal component and $\alpha_t$, $\beta_t \in [0, 1]$ are smoothing coefficients. Note that in contrast to the standard ES, the smoothing coefficients $\alpha_t$, $\beta_t$ change with time $t$. They are calculated from the RNN outputs, $\Delta\alpha_t$ and $\Delta\beta_t$:
\begin{equation}
\begin{aligned}
\alpha_{t+1} = \sigma(\alpha_{0} + \Delta\alpha_t), \\
\beta_{t+1} = \sigma(\beta_{0} + \Delta\beta_t).
\label{eqab}
\end{aligned}
\end{equation} 
Here $\alpha_{0}$, $\beta_{0}$ are initial values of the smoothing coefficients and $\sigma$ is a sigmoid function, which maintains the coefficients within $(0,1)$ range.

\textbf{Neural network backbone} is a deep residual RNN, which we call RNN for simplicity. RNN has several inputs, including the randomly generated quantile level $q$ and a normalized, deseasonalized vector of the past $w=168$ values:
\begin{equation}
x_\tau=\log{\frac{y_\tau}{\bar{y}_t \widehat{s}_{t,\tau}} }\,,
\label{eqxt}
\end{equation}
where $\tau \in \Delta^{in}_t$, $\Delta^{in}_t$ denotes the input rolling window, which is adjacent to the output window $\Delta^{out}_t$ of length $H=48$ covering the forecast horizon, $\bar{y}_t$ is the average value in $\Delta^{in}_t$ and $\widehat{s}_{t,\tau}$ is the seasonal component produced by ES. 
 
The input vector concatenates $q$ and $\textbf{x}'_t = [x_\tau]_{\tau \in \Delta^{in}_t}$ with: (i) date features, $\textbf{d}_t^{w} \in \{0, 1\}^7, \textbf{d}_t^{m} \in \{0, 1\}^{31}$ and $\textbf{d}_t^{y} \in \{0, 1\}^{52}$, \emph{i.e.} binary one-hot vectors encoding day of the week, day of the month and week of the year, respectively; (ii) seasonality components, $\widehat{\textbf{s}}_t$, \emph{i.e.} a vector of 48 seasonal components predicted by ES for the output period $t$ reduced by 1 and (iii) logarithm of the normalizer, $\log_{10}(\bar{y}_t)$. 

The input is fed to the deep dilated RNN architecture consisting of 5 recurrent layers as shown in Fig.~\ref{fig:AQESRNN}. RNN employs attentive dilated RNN cells, adRNNCells, introduced in \cite{smyl2022dyn}. They possess a receptive field whose size is determined by a dilation parameter ($d \geq 1$). Moreover, they incorporate an attention mechanism that dynamically adjusts the weighting of input information, making them particularly effective at capturing complex dynamics and seasonalities. Larger dilations are assigned to top layers, \emph{i.e.} $d=2, 4, 7, 14, 28$. The dilations are in day units, as we step in time by 24 hours. Residual connections flowing around each adRNNCell are added to avoid vanishing gradient problem. Finally, an important feature of the architecture is that the input containing a reference to the quantile probability $q$ is repeatedly appended to the input of each layer, which improves the distributional forecasting accuracy according to our ablation studies.

\textbf{Postprocessing}. The postprocessing component converts the RNN output, represented as a vector of 
$H=48$ normalized and deseasonalized $q$-th quantiles, $\widehat{\mathbf{y}}'_{q,t}$, into actual quantiles, $\widehat{\mathbf{y}}_{q,t}$, using inverted equation \eqref{eqxt}:
\begin{equation}
\widehat{y}_{q,\tau}=\exp{(\widehat{y}'_{q,\tau})}
\bar{y}_t \widehat{s}_{t,\tau},\quad \tau \in \Delta^{out}_t \,.
\label{eqn:unknown}
\end{equation}
We also found empirically that postprocessing the RNN forecast via quantile sorting noticeably improves distributional forecasting accuracy. For every horizon length (1 to 48) we sort the forecasted values for all quantiles and then match them with the sorted quantile probabilities. This prevents the quantile crossing and greatly improves the forecast accuracy (please refer to ablation studies for details).

\begin{figure}[t]
    \centering
    \includegraphics[width=0.5\linewidth]{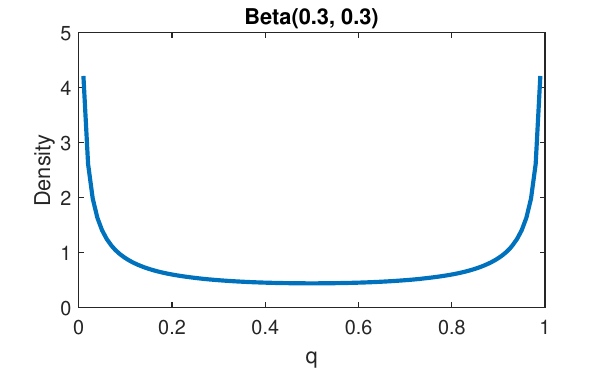}
    \caption{Beta(0.3, 0.3) distribution used for the training of AQ-ESRNN.}
    \label{fig:beta}
\end{figure}

\textbf{Training Methodology}. We use standard quantile loss function~\eqref{eqn:quantile_loss}, operating on normalized values, \emph{i.e.} $y^n=y/\bar{y}$ and $\widehat{y}^n_q =\widehat{y}_q/\bar{y}$. A unique quantile probability $q$ is sampled randomly for each time-series in the batch. It is worth mentioning that we found that AQ-ESRNN training benefits from non-uniform sampling of $q$ with emphasis on the tails. We attribute this to the fact that ESRNN uses saturating activation functions, which lead to diminishing gradients at the tail of the distribution. Thus it is more challenging to fit well to very small (close to 0) and very large (close to 1) quantiles, because the quantile function is likely to change quickly in those regions. We conjecture that this can be compensated for by denser sampling of quantile probabilities in the tails. In particular, we use symmetric Beta distribution with both shape parameters less than 1, \emph{e.g.}, Beta(0.3, 0.3) (see Fig.~\ref{fig:beta}). 

\subsection{AQ-NBEATS} \label{ssec:aq_nbeats}

\begin{figure}[t]
    \centering
    \begin{subfigure}[t]{0.45\textwidth}
        % \centering
        \includegraphics[width=\textwidth]{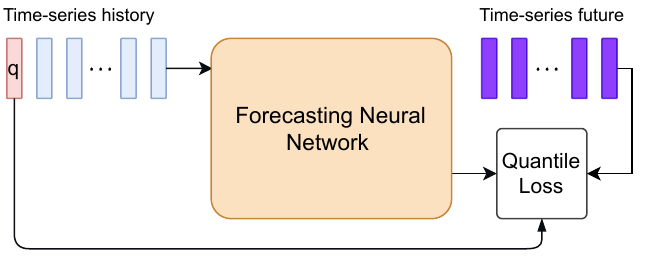}
        \caption{AQ-CAT, the target quantile probability is concatenated as one of the elements in the input sequence}\label{fig:any_quantile_variants:aqcat}
    \end{subfigure}
    \hspace{0.08\textwidth}
    \begin{subfigure}[t]{0.45\textwidth}
        % \centering
        \includegraphics[width=\textwidth]{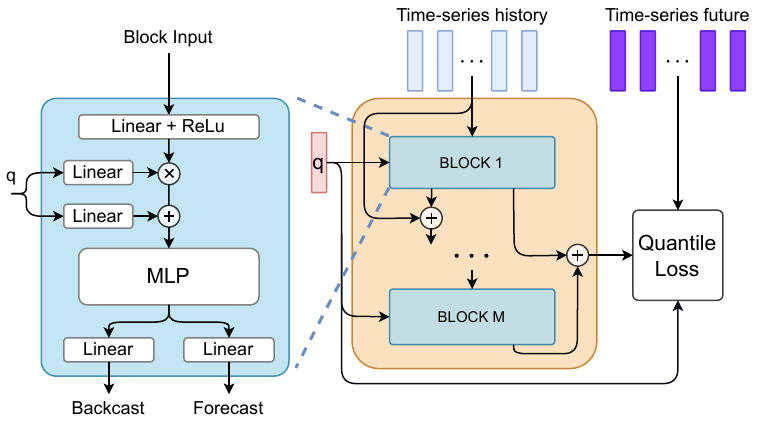}
        \caption{AQ-FiLM, the target quantile probability is injected in every N-BEATS block via FiLM modulation layer.}\label{fig:any_quantile_variants:aqFiLM}
    \end{subfigure}
    \begin{subfigure}[t]{0.45\textwidth}
        % \centering
        \includegraphics[width=\textwidth]{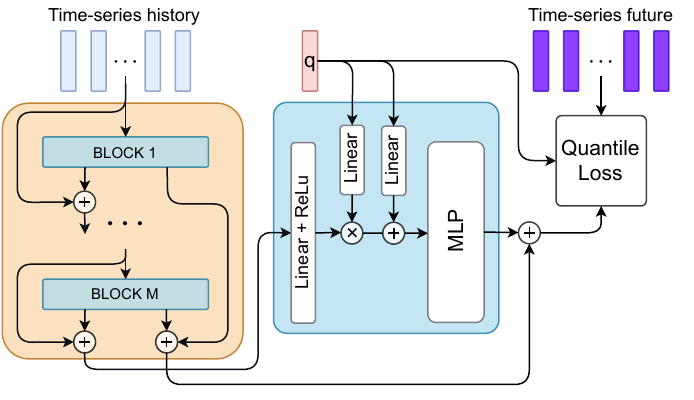}
        \caption{AQ-OUT, the target quantile probability is injected in the last N-BEATS block only.}\label{fig:any_quantile_variants:aqout}
    \end{subfigure}
\caption{Variants of implementing the any-quantile conditioning in N-BEATS} 
\label{fig:any_quantile_variants} 
\end{figure}

Our exposition of the original N-BEATS architecture closely follows~\cite{oreshkin2021nbeats}. Each block of N-BEATS is a residual architecture consisting of the sequence of fully connected blocks with forecast/backcast forks at the end. The architecture runs a residual recursion over the entire input window and sums block outputs to make its final forecast. We assume that there are $R$ residual blocks each having $L$ hidden layers. If we refer to $\nbeatsinput \in \mathbb{R}^{\windowlength}$ as the input of the architecture of length $\windowlength$, use residual block and layer superscripts ($r$ and $\ell$, respectively) and denote the fully connected layer with weights $\vec{W}^{r,\ell}$ and biases $\vec{b}^{r,\ell}$ as $\fc_{r,\ell}(\nbeatshidden^{r, \ell-1}) \equiv \relu(\vec{W}^{r,\ell} \nbeatshidden^{r,\ell-1} + \vec{b}^{r,\ell})$, the operation of N-BEATS is described as follows:
\begin{align}  \label{eqn:nbeats_fc_network}
\begin{split}
    \vec{x}^{r} &= \nbeatsinput^{r-1} - \nbeatsbackcast^{r-1}, \\
    \nbeatshidden^{r,1} &= \fc_{r,1}(\vec{x}^{r}), \  \ldots, \  \nbeatshidden^{r,L} = \fc_{r,L}(\nbeatshidden^{r,L-1}),  \\
    \nbeatsbackcast^{r} &= \vec{B}^{r} \nbeatshidden^{r,L}, \    \nbeatsforecast^{r} = \vec{F}^{r} \nbeatshidden^{r,L}.
\end{split}
\end{align}
Here we have $\nbeatsbackcast^{0} \equiv \vec{0}$, $\nbeatsinput^{0} \equiv \nbeatsinput$ and projection matrices have dimensions $\vec{B}^{r} \in \mathbb{R}^{\windowlength \times d_h}$, $\vec{F}^{r} \in \mathbb{R}^{H \times d_h}$. Internal hidden dimension of each block is $d_h$. The final forecast is the sum of forecasts of all the residual blocks, $\nbeatsforecast = \sum_r \nbeatsforecast^{r}$. Given the structure of the network, there are three principal ways, depicted in Fig.~\ref{fig:any_quantile_variants} and described in detail below, that can be used to feed quantile $\vec{q}$ to N-BEATS. 

\textbf{Input quantile concatenation scheme} is the easiest (does not require architectural modifications) and the most generic approach that is applicable to N-BEATS as well as any other forecasting architecture. In this scheme, called AQ-NBEATS-CAT and depicted in Figure~\ref{fig:any_quantile_variants:aqcat}, quantile level is simply appended as one of the elements (\emph{e.g.} the last element) in the input time-series history. The caveat here is that in the univariate forecasting scenario the original input has dimensions $B \times \windowlength$ (batch size times the history length). In order to enable quantile computation for multiple quantiles $Q$ for each time-series, the quantile-appended input will now have the form of 3D tensor of the size $B \times Q \times (\windowlength+1)$, which multiplies the total computational cost of predicting $Q$ quantiles at inference time by a factor of $Q$. Additionally, quantile level is restricted to the range $[0, 1]$. Hence, concatenating it with inputs of large magnitude can create dynamic range issues and make the network completely ignore the $\vec{q}$ input. Therefore, additional input normalization measures may have to be necessary to ensure effective quantile operation.

\textbf{Block quantile conditioning scheme} depicted in Figure~\ref{fig:any_quantile_variants:aqFiLM} is more specific to N-BEATS and implies architectural modification. In this scheme, which we call AQ-NBEATS-FiLM, we propose to use the FiLM conditioning method~\cite{perez2018FiLM} to modulate the computation of each N-BEATS block with the representation derived from the quantile probability. In terms of inference path, the first and the last equation of AQ-NBEATS-FiLM block replicate original N-BEATS~\eqref{eqn:nbeats_fc_network}, whereas the middle part is modified as follows:
\begin{align}  \label{eqn:aqnbeats_fc_network}
\begin{split}
    \nbeatshidden^{r,1} &= \fc_{r,1}(\vec{x}^{r})\,, \\
    \vec{c}^{r}_{\alpha} &= \linear_{r, \alpha}(\vec{q}), \quad \vec{c}^{r}_{\gamma} = \linear_{r, \gamma}(\vec{q})\,, \\
    % \vec{c}^{r}_{\alpha} &= \vec{W}^{r,\alpha} q, \quad \vec{c}^{r}_{\gamma} = \vec{W}^{r,\gamma} q \\
    \nbeatshidden^{r,2} &= \vec{c}^{r}_{\alpha} + (1 + \vec{c}^{r}_{\gamma}) \nbeatshidden^{r,1}\,, \\
    &\ldots  \\
    \nbeatshidden^{r,L} &= \fc_{r,L}(\nbeatshidden^{r,L-1})\,.
\end{split}
\end{align}
FiLM induces both additive shift and multiplicative modulation of the internal time-series representation based on the target quantile probability. Since both multiplicative and additive components are included, this scheme is less sensitive to the issues that may arise from the lack of normalization, the problem we pointed out in relation with the AQ-NBEATS-CAT variant. Additionally, we hypothesise that the inclusion of quantile conditioning in each block should lead to better overall modeling of quantiles, since the architecture is effectively ``reminded'' of the quantile forecasting task in every block, whereas in AQ-NBEATS-CAT, the quantile hint is provided only once at the input and by the time the representation reaches the last block of a deep network it can effectively be washed away.

\textbf{Output quantile conditioning scheme} called AQ-NBEATS-OUT is depicted in Figure~\ref{fig:any_quantile_variants:aqout}. Here the bulk of the processing is done in a standard forecasting architecture, while quantile conditioning is done in the special output quantile conditioning block. This is a more computationally effective variant of AQ-NBEATS-FiLM. In AQ-NBEATS-FiLM, there is a computational cost for constantly ``reminding'' the architecture about the quantile it has to produce. Since quantile conditioning happens in every block, the architecture persistently works with $B \times Q \times d_h$ tensors ($d_h$ is the dimensionality of hidden layers), multiplying the total computational cost of predicting $Q$ quantiles at inference time by a factor of $Q$. We hypothesize that the bulk of the work of representing a time-series can be done without quantile conditioning. Therefore, if we apply quantile conditioning only in the last block, as depicted in Figure~\ref{fig:any_quantile_variants:aqout}, we should be able to obtain comparable accuracy results while significantly saving on computation. For example, with a typical configuration of N-BEATS depth of 10-30 blocks, the computational cost of AQ-NBEATS-OUT will be 10-30 times smaller than that of AQ-NBEATS-FiLM.

\textbf{Training Methodology}. To train AQ-NBEATS, we use standard quantile loss function~\eqref{eqn:quantile_loss}. A unique quantile probability $q$ is sampled for each time-series in the batch, uniformly at random.

% \subsection{Any-quantile training procedure} \label{ssec:aq_training_procedure}

% The quantile level is generated uniformly at random in the $[0,1]$ range for each training sample and is used both as neural network input and as the supervision signal in the quantile loss. As a result of this training, the neural network learns to generate forecast for any desired quantile level conditional on historical inputs. 

% \subsection{AQ-Transformer}

% \subsection{AQ-CNN}

\section{Empirical Results} \label{sec:results}

In this section, we conduct a comprehensive assessment of our proposed AQ-models in the context of short-term electricity demand forecasting across 35 European countries. The evaluation process unfolds as follows: we provide insights into the nature of the data, elucidate the methodologies involved in training and evaluation and describe models used as baselines. Subsequently, we present the results, delve into an ablation studies and conclude by engaging in a thorough discussion regarding the distinctive properties of our models.

\subsection{Datasets}

\begin{figure}[t]
    \centering
    \includegraphics[width=0.5\linewidth]{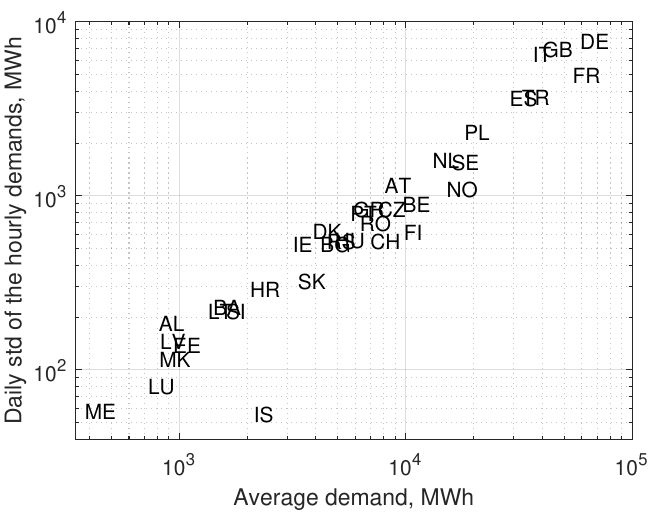}
    \caption{Average electricity demand and daily dispersion demand in 35 European countries.}
    \label{fig:AvStd}
\end{figure}

The real-world dataset utilized in this study is sourced from the ENTSO-E repository (\url{www.entsoe.eu/data/power-stats}). This dataset encompasses hourly electricity demand data for 35 European countries and we have made this dataset publicly available along with the model code in our GitHub repository 
(To be released if the paper is accepted).
%\url{https://github.com/boreshkinai/any-quantile}. 
The dataset contains a rich collection of time-series, each possessing distinct characteristics such as level, trend and variance varying over time, seasonalities (annual, weekly and daily) and random fluctuations of varying intensity. The inclusion of time-series with such diverse properties ensures a robust evaluation of the model's performance.

Fig. \ref{fig:AvStd} provides a snapshot of the diverse energy consumption profiles across European countries. It illustrates the average demand and daily dispersion of hourly demands over the three-year period from 2016 to 2018. The average demand exhibits notable variability, ranging from 383 MWh for Montenegro (ME) to 58,941 MWh for Germany (DE). Similarly, the daily standard deviation of hourly demands varies from 56 MWh (Iceland, IS) to 7,687 MWh (DE). 

\begin{figure}[t]
    \centering
    \includegraphics[width=0.7\linewidth]{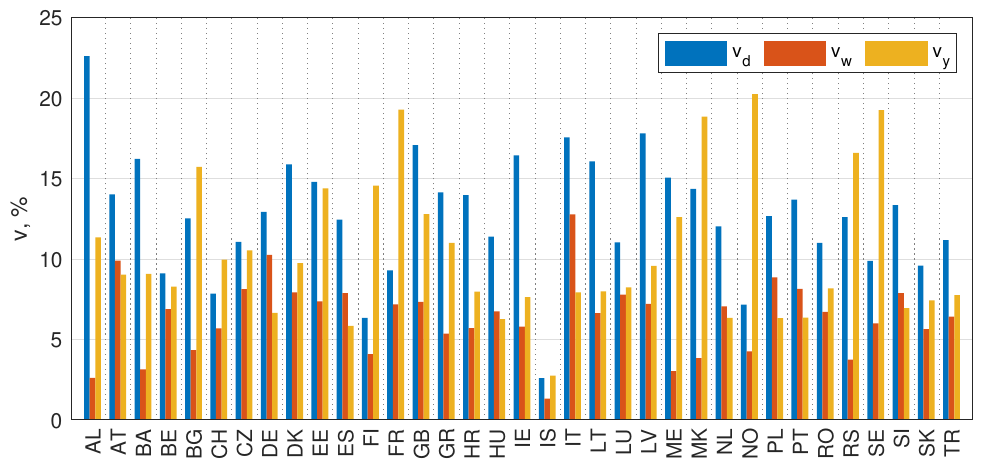}
    \caption{Diversity of electricity demand in 35 European countries.}
    \label{fig:Countries}
\end{figure}

The variability in demand significantly influences the width of the probabilistic forecast distribution. We quantify this variability as the ratio of the standard deviation to average demand value: $v=100s/\bar{z}$. We consider it in three contexts: as daily demand variations ($v_d$), where $\bar{z}$ and $s$ denote the daily mean and standard deviation, respectively; as weekly variations ($v_w$), where $\bar{z}$ and $s$ represent the weekly mean and standard deviation of daily means; and as yearly variations ($v_y$), where $\bar{z}$ and $s$ express the yearly mean and standard deviation of weekly means. Figure~\ref{fig:Countries} shows the variation coefficients for 35 European countries. Notably, Iceland exhibits the least demand variation. The most pronounced daily variations ($v_d>17\%$) are observed in Albania, Italy, Latvia and Great Britain. For yearly variations, Norway, France, Sweden and Macedonia demonstrate the highest variability ($v_y>18\%$). Weekly variations are generally milder than daily and yearly ones, with $v_w<13\%$. Italy, Germany, Austria and Poland show the strongest weekly variations.
It is crucial to highlight that the pattern of electricity demand variation evolves over time, presenting additional challenge for forecasting models.

\subsection{Training and Evaluation Setup}

The dataset is split into train, validation and test subsets with the following boundaries: train dates 2006-01-01 to 2016-12-31, validation dates 2017-01-01 to 2018-12-31, test dates 2018-01-01 to 2018-12-31. In the training set, NaN (missing) values are forward filled (replacing the missing value with a value of 168 hours, a full week, earlier time) so that they can be utilized for training. In the test set, NaN values are excluded from metric calculations to avoid any information leakage. We use the train and validation subsets to tune hyperparameters. Once the hyperparameters are determined, the model with best hyperparameter settings, re-trained on the train and validation sets, is used to report results on the test set. The prediction horizon is 48, or two days.
A prediction horizon of this length is imposed by the requirements of power system operation. The system operator typically prepares forecasts for the next day well in advance, for example 12 hours in advance according to~\cite{Pop15}. For final evaluation we employ the ensemble of 8 models, each sharing the same structure but initialized randomly and exposed to distinct, randomized sequences of input data. The final forecasts are generated by taking the medians of all ensemble forecasts.

\subsection{AQ-ESRNN Training and Hyperparameter Settings} \label{ssec:aq_esrnn}

\begin{table*}[!t]
    \centering
    \caption{Settings of AQ-ESRNN hyperparameters and the hyperparameter search grid.}
    \label{table:esrnn_hyperparameter_settings}

    \begin{tabular}{lcc}
        \toprule
        Hyperparameter & Value  & Grid \\ 
        \midrule
        Batch size & 2, changed to 5 in ep. 4 & changed to 5 in ep. 3, 4\\ 
        Epochs	& 6 & 4, ..., 9	\\ 
        Learning rates & 0.003, /3 in ep. 5, /10 in ep. 6 & also /30 in ep. 7\\        
        LR multiplier & 30 & [10, 30]  \\
        Beta & 0.3 & [0.1,0.2,0.3,0.5,0.65,0.8,1]	\\
        Dilations & [[2],[4],[7],[14],[28]] & see Section~\ref{ssec:aq_esrnn}  \\
        training steps & 50 & [20, 30, 50]  \\
        c-state size & 150 & [100, 150, 200] \\
        h-state size & 70 &  [50, 70, 100] \\
        $\alpha_0$  & 3.5 & --- \\
        $\beta_0$ & 0.3 & --- \\
        \bottomrule
    \end{tabular}
\end{table*}

All AQ-ESRNN hyperparameters are adjusted by minimizing $\crps$ on the validation set. The objective function used to train the network is quantile loss~\eqref{eqn:quantile_loss}. The model is implemented in PyTorch~\cite{paszke2019pytorch} and is trained for 6 epochs corresponding to around 1500 batch updates using Adam optimizer~\cite{kingma2015adam}. We use a schedule of increasing batch sizes. Initially, we use batch size of 2 corresponding to 1750 updates per epoch and as batch size is increased to 5, the number of batch updates decreases to approximately 1450 per epoch. In addition to batch size scheduling, we use learning rate scheduling. On top of this, model contains global learnable parameters and time-series specific learnable parameters. For the latter, there is LR multiplier boosting the learning rate for per-series parameters. When experimenting with models having fewer layers, we observed earlier signs of overfitting. Therefore, for 3 and 4-layer models, we report results from epoch 5, while for 5-layer models, we report results from epoch 6. The training of a full model with 5 layers takes around 2 days on CPU.

Table~\ref{table:esrnn_hyperparameter_settings} summarizes optimal hyperparameter settings of the AQ-ESRNN model, alongside the ranges explored during hyperparameter tuning process. A few additional remarks are in order to explain the fields of the table. \emph{Beta} is a parameter utilized for a symmetrical Beta distribution, which is employed in generating the quantiles during training. \emph{Dilations} describe the architecture of AQ-ESRNN. Each internal square bracket lists the dilations of layers in a particular block. For instance, [[2], [4], [7], [14], [28]] describes a system composed of 5 blocks, each with a single layer and a progressively increasing delay of 2, 4, 7, 14 and 28 time steps. In experiments with 3-layer architecture, dilations are set as [[2], [4], [7]]. For 4-layer architecture, we experimented with dilations like [[2], [4], [7], [14]], [[2], [4], [7], [7]] and [[2], [4], [7], [4]]. Similarly, for 5-layer architectures, we tested dilations such as [[2], [4], [7], [14], [28]], [[2], [4], [7], [14], [14]], [[2], [4], [7], [14], [7]] and [[2], [4], [7], [10], [14]].
\emph{Training steps} -- when a batch is formed from randomly chosen series, the RNN time steps initiate from a randomly selected step and persist for this number of steps. \emph{C-state size} and \emph{h-state size} -- the cells employed are derived from LSTMs, but differ in several aspects, including c-state and h-state having different sizes. The c-state size must be larger than h-state size, because the cell output size equals c-state size minus h-state size.

\subsection{AQ-NBEATS Training and Hyperparameter Settings} 

AQ-NBEATS model is evaluated using the hyperparameter settings presented in Table~\ref{table:nbeats_hyperparameter_settings}. Similarly to AQ-ESRNN, the hyperparameters are adjusted by minimizing $\crps$ on the validation set. The objective function used to train the network is normalized quantile loss:
\begin{equation}
\rho_{N}(y, \widehat{y}_q) =
\begin{cases}
(y-\widehat{y}_q)q / y\,,       & \text{if } y \geq \widehat{y}_q \,,\\
(y-\widehat{y}_q)(q-1) / y\,,  &\text{if } y < \widehat{y}_q \,.
\end{cases}
\label{eqn:mqn-loss-definition}
\end{equation}
The model is implemented in PyTorch~\cite{paszke2019pytorch} using pytorch lightning framework~\cite{falcon2019pytorch} trained for 15 epochs using the Adam optimizer~\cite{kingma2015adam} with default settings using inverse square root learning rate annealing with maximum learning rate 0.0005 and warm-up period of 400 batches. Batch has size 1024 and the model takes the history of 168 points (24 hours times 7 days) and predicts 48 points (two times 24 hours) ahead in one shot. 

\begin{table*}[!t]
    \centering
    \caption{Settings of AQ-NBEATS hyperparameters and the hyperparameter search grid.}
    \label{table:nbeats_hyperparameter_settings}

    \begin{tabular}{lcc}
        \toprule
        Hyperparameter & Value  & Grid \\ 
        \midrule
        Epochs	& 15 & [10, 15, 20]	\\ 
        Loss & Quantile,~\eqref{eqn:mqn-loss-definition} & [\eqref{eqn:mqn-loss-definition}, \eqref{eqn:quantile_loss}] \\
        Input Normalization & No & [Yes, No] \\
        Width ($d_h$) & 1024 & [256, 512, 1024]	  \\
        Blocks ($R$) & 30 & [5, 10, 20, 30]	  \\
        Layers ($L$) & 3 & [2, 3, 4]  \\
        Sharing & False & [True, False]	  \\
        Lookback period ($w$, hours) & 168 & [24, 168, 336]  \\
        Batch size & 1024 & [256, 512, 1024]	\\
        Optimizer & Adam & Adam \\
        Learning rate & 0.0005 & [0.001, 0.0005]\\
        \bottomrule
    \end{tabular}
\end{table*}

\subsection{Baseline Models}

We compare our AQ-models against a number of probabilistic forecasting baselines, encompassing both statistical and neural network models. First, we list the baselines and in the ensuing paragraphs we explain important implementation details. \textbf{Naive}, a seasonal naive method: the forecasted demand profile for day $t$ is the same as the profile for day $t-7$. \textbf{ARIMA}, autoregressive integrated moving average model~\cite{Dud15}. \textbf{ES}, exponential smoothing model~\cite{Dud15}. \textbf{Theta}, dynamic optimised Theta model, DOTM v3 from~\cite{Dud19}. \textbf{MLP}, single-hidden-layer perceptron with sigmoid nonlinearities~\cite{Dud16}. \textbf{DeepAR}, autoregressive RNN model for probabilistic forecasting~\cite{Sal20}. \textbf{Transformer}, vanilla transformer~\cite{Vas17}. \textbf{WaveNet}, deep autoregressive model combining causal filters with
dilated convolutions~\cite{Oor16}. \textbf{TFT}, temporal fusion transformer \cite{ Lim21}.

Naive and statistical models generate point forecasts.
To deal with triple seasonality in electricity demand time-series, we decompose these series into 24 distinct series, each representing a particular hour of the day. For each of these 24 series, a separate statistical model is utilized to forecast electricity demand, with a horizon of two. Consequently, all models produce a total of 48 forecasts for the subsequent two days. For Naive, ARIMA and ES, we assume that the distribution of possible future values follows a normal distribution. The point forecast represents the mean of this distribution and the standard deviation is estimated using the standard deviation of residuals \cite{Hyn21}.
To obtain quantiles, the models are first fitted using their respective R functions in the \texttt{forecast} package: \texttt{snaive()}, \texttt{auto.arima()} and \texttt{ets()}. Quantiles are then derived by setting the \texttt{level} parameter in the \texttt{forecast()} function to specific value defining PI. In contrast, the Theta method estimates quantiles through bootstrapping, which involves simulating a sample of possible values of forecasts from the estimated model \cite{Fio16}. This technique is implemented in the \texttt{dotm()} function in the \texttt{forecTheta} package.

Neural models employed in this study are sourced from GluonTS, a Python package designed for probabilistic time-series modeling, with a primary focus on deep learning models \cite{Ale20}. Specifically, the MLP, DeepAR and Transformer models are utilized to generate probabilistic forecasts through a parametric approach. These models have a projection layer that produces distribution parameters. Model fitting process involves the negative log-likelihood of the chosen distribution as the loss function. We use Student’s t-distribution, which is the default setting. Once the model is fitted, quantiles can be derived from the learned paramteric distribution. In contrast, WaveNET models the auto-regressive forecast probability distribution by outputting a softmax-induced categorical distribution over a fixed grid of quantiles. Finally, TFT simultaneously predicts multiple quantiles at each time step alongside a point forecast. Quantile forecasts are derived by applying a linear transformation on top of the TFT decoder embedding. During training, TFT learns by jointly minimizing the quantile loss aggregated across all quantile outputs. Arbitrary quantiles are derived from the fixed quantile grid via interpolation.

\begin{table*}[!t]
\centering
\caption{Key performance metrics. Lower score is better. Best result is shown in bold.}
\begin{tabularx}{\textwidth}{l@{\extracolsep{\fill}} ccc}
 & CRPS & N-CRPS & MAPE  \\
\midrule
Naive       & 502.62                            & 3.95                                & 5.08                              \\
ARIMA       & 353.49                            & 2.83                                & 3.74                              \\
ES          & 325.80                            & 2.64                                & 3.48                              \\
Theta       & 329.99                            & 2.69                                & 3.56                              \\
\midrule
MLP         & 415.67                            & 3.28                                & 4.49                              \\
DeepAR      & 378.46                            & 2.97                                & 3.95                              \\
WaveNet     & 293.38                            & 2.52                                & 3.39                              \\
Transformer & 380.41                            & 2.83                                & 3.72                              \\
TFT         & 325.80                            & 2.92                                & 3.96                              \\
\midrule
AQ-ESRNN    & \textbf{195.94}                   & \textbf{1.72}                       & \textbf{2.32}                     \\
AQ-NBEATS   & 211.22 & 1.84 & 2.47  \\
\bottomrule
\end{tabularx}%
\label{table:key_results}
\end{table*}%

\subsection{Key Quantitative Results}

Our key results are presented in Table~\ref{table:key_results}. They include CRPS~\eqref{eqn:crps_l2}, normalized CRPS (N-CRPS) and MAPE for assessing point forecasts derived from distributional forecasts. Normalized version of CRPS is defined as follows:
\begin{align}
\ncrps(f_{\theta}, \vec{y}, \vec{x}) = 100 \frac{1}{C} \sum_{c} \frac{1}{H N_c Q} \sum_{i,h,q} \rho(y_{i,h,c}, f_{\theta}(\vec{x}_{i,c}, q)) / \bar{y}_{c} \,.
\label{eqn:ncrps}
\end{align}
Here $C=35$ is the number of countries, $H=48$ is the number of horizons, $Q=100$ is the number of random test quantiles and $N_c$ is the number of test samples per country. The second sum runs over samples in the country time series, horizons and quantiles. $\bar{y}_c$ is the mean of a given country time-series. Normalized CRPS unifies results for different countries by normalizing over mean demand level $\bar{y}_c$. Therefore, all countries can be compared with each other and their impact in the average N-CRPS value shown in Table~\ref{table:key_results} is similar.

We see that the any-quantile forecasting task is challenging for classical time-series models as well as for the out-of-the-box models from GluonTS. 
Among statistical models, ES emerges as top performer, while WaveNet leads the neural baselines. WaveNet slightly outperforms ES. Our novel any-quantile architectures, AQ-ESRNN and AQ-NBEATS, set new benchmarks in distributional forecasting, achieving the lowest CRPS. Additionally, when comparing mean absolute percentage errors (MAPE) as the performance measure for point forecasting, where the 0.5-quantile (median) is treated as a point forecast, the AQ-models surpass all baselines. Notably, AQ-ESRNN slightly outperforms AQ-NBEATS across both CRPS and MAPE metrics.

\begin{figure}[t]
    \centering
    \includegraphics[width=0.50\linewidth]{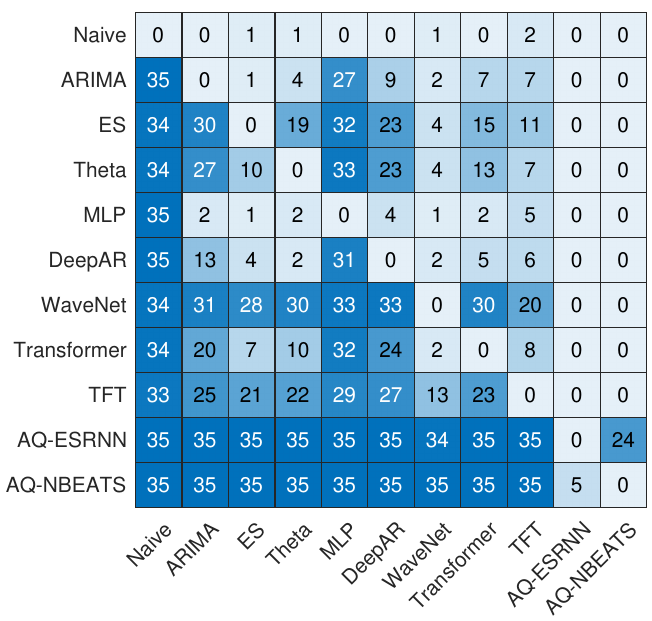}
    \caption{Results of the Diebolt-Mariano test based on the N-CRPS metric. Individual cells count the number of countries the model represented on the y-axis is statistically more accurate than the model represented on the x-axis.}
    \label{fig:DM}
\end{figure}

To assess the statistical significance of differences between the probabilistic forecasts generated by each pair of models, we conduct a Diebold-Mariano test \cite{Die95}, with respect to the individual country errors. The results, based on the N-CRPS metric are shown in Fig. \ref{fig:DM}, which counts the number of countries the model represented on the y-axis is statistically more accurate than the model represented on the x-axis. It is noteworthy that both our AQ-models demonstrate superior accuracy over baseline models for nearly all countries. Furthermore, AQ-ESRNN exhibits higher accuracy than AQ-NBEATS in 24 countries, while AQ-NBEATS outperforms AQ-ESRNN in 5 countries.

\begin{figure}[t]
    \centering
    \includegraphics[width=0.48\linewidth]{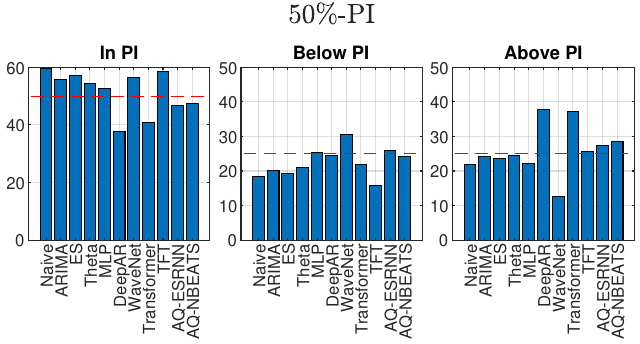}
    \includegraphics[width=0.48\linewidth]{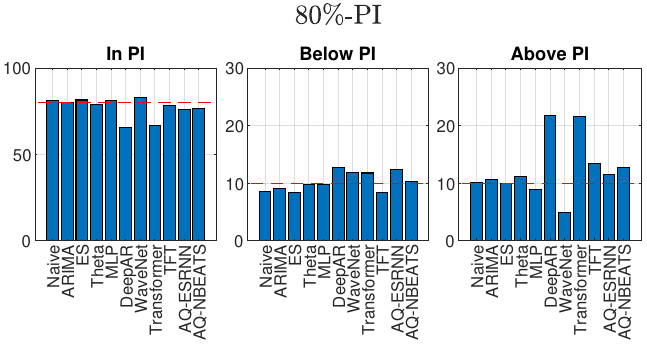}
    \includegraphics[width=0.48\linewidth]{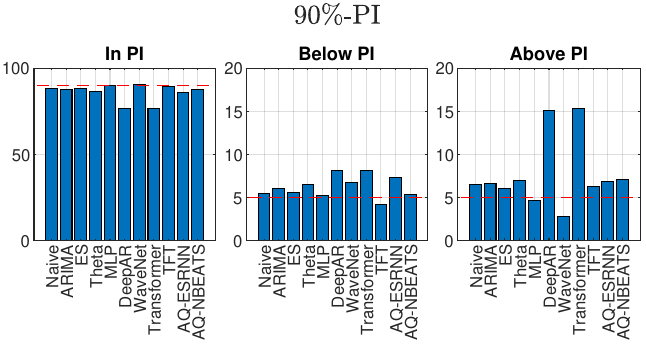}
    \includegraphics[width=0.48\linewidth]{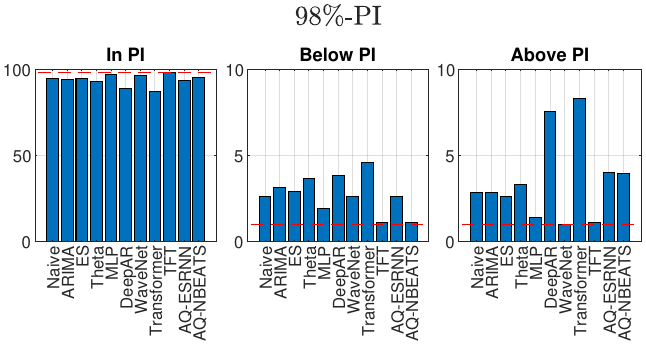}
    \includegraphics[width=0.48\linewidth]{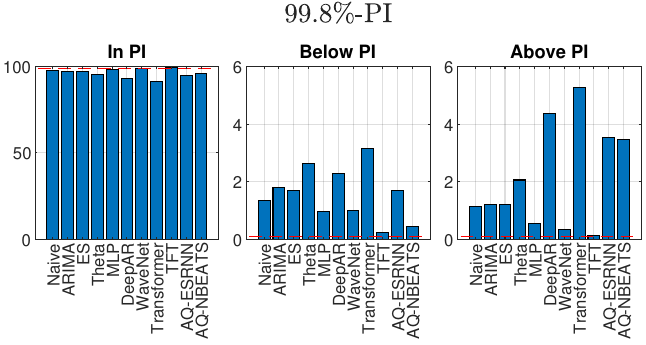}
    \caption{Percentage of forecasts in, below and above the PI of different size. Red dashed line indicates the target value.}
    \label{fig:Div}
\end{figure}

Figure \ref{fig:Div} presents the evaluation of PIs of varying sizes generated by the forecasting models. Bars indicate the proportion of observed values that fall within, below, or above the boundaries of the PIs predicted by the models, while the red dashed lines signify the theoretical target (for instance, for a 50\%-PI, the targets are 50\% within, 25\% below and 25\% above the PIs). In the context of In-PI charts, bars surpassing the target level signify PIs that are excessively broad, implying an overestimation of the uncertainty range. Conversely, bars falling short of target  suggest PIs that are unduly narrow, indicating an underestimation of uncertainty. The pattern of too narrow PIs is particularly evident in the cases of the DeepAR and Transformer models, whereas an inclination towards too broad PIs is primarily observed for 50\%-PI in the statistical models, Naive, WaveNet and TFT. When examining the Below-PI and Above-PI charts, the most striking observation is that for DeepAR and Transformer the upper PI bounds are notably too low, with the bars in the Above-PI charts significantly above the target values. Conversely, WaveNet exhibits an opposite trend, with excessively high upper PI bounds. For broader PIs, specifically those at 98\% and 99.8\%, predicted PIs are often too tight, indicating challenges in forecasting extreme quantiles. In these scenarios, the TFT model exhibits superior performance. Excluding these extreme quantiles, the performance of our any-quantile models stands out, achieving PIs that closely approximate the targets.

%Makridakis, S., Spiliotis, E., Assimakopoulos, V., Chen, Z., Gaba, A., Tsetlin, I., Winkler, R.L.: The M5 uncertainty competition: Results, findings and conclusions. International Journal of Forecasting 38(4), 1365–1385 (2022)

%Gneiting, T., & Raftery, A. E. (2007). Strictly proper scoring rules, prediction and estimation. Journal of the American Statistical Association, 102(477), 359–378. http://dx.doi.org/10.1198/016214506000001437.

\begin{figure}[t]
    \centering
    \includegraphics[width=0.24\linewidth]{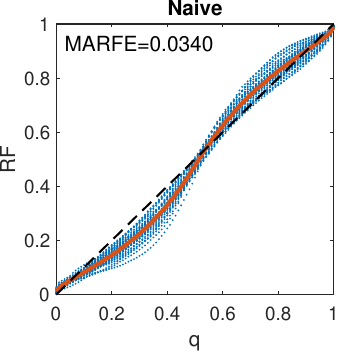}
    \includegraphics[width=0.24\linewidth]{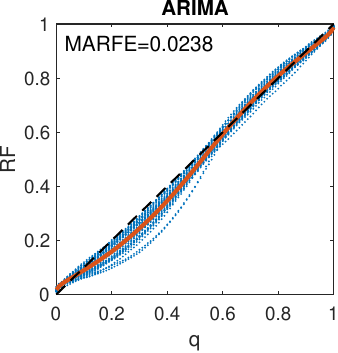}
    \includegraphics[width=0.24\linewidth]{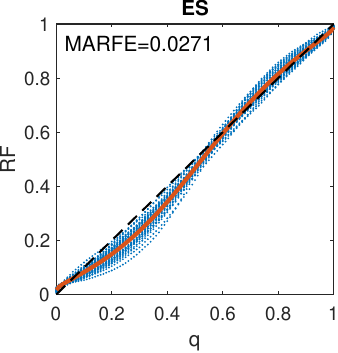}
    \includegraphics[width=0.24\linewidth]{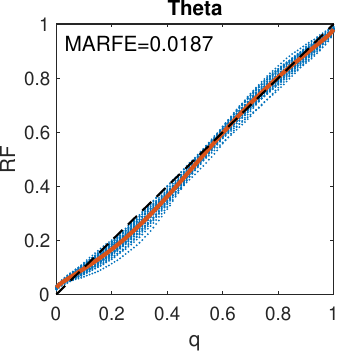}
    \includegraphics[width=0.24\linewidth]{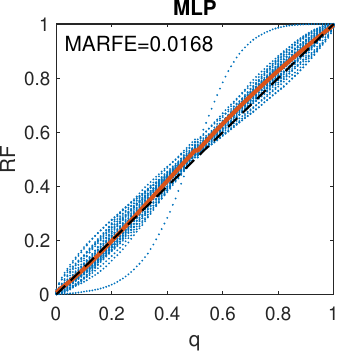}
    \includegraphics[width=0.24\linewidth]{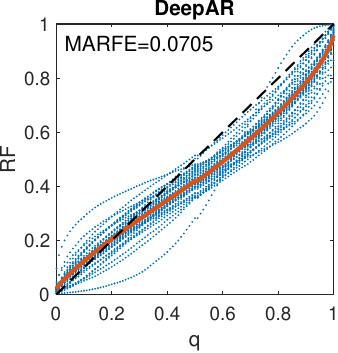}
    \includegraphics[width=0.24\linewidth]{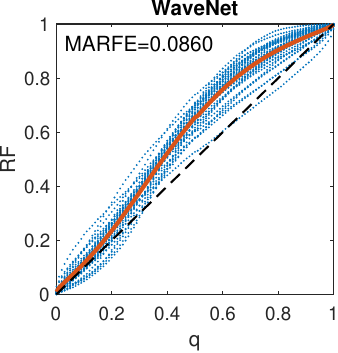}
    \includegraphics[width=0.24\linewidth]{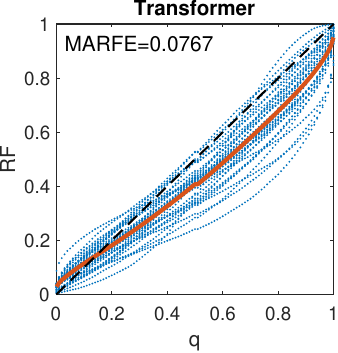}
    \includegraphics[width=0.24\linewidth]{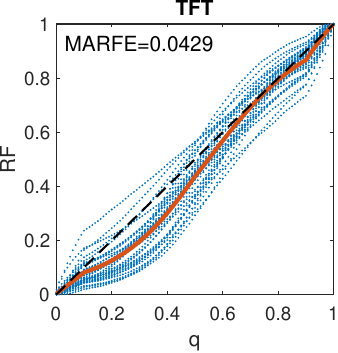}
    \includegraphics[width=0.24\linewidth]{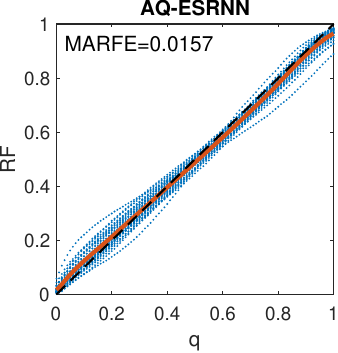}
    \includegraphics[width=0.24\linewidth]{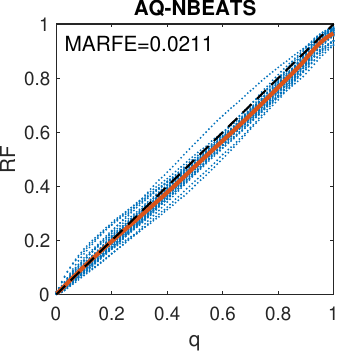}
    \caption{Relative frequency.}
    \label{fig:RF}
\end{figure}

\begin{figure}[t]
    \centering
    \includegraphics[width=0.248\linewidth]{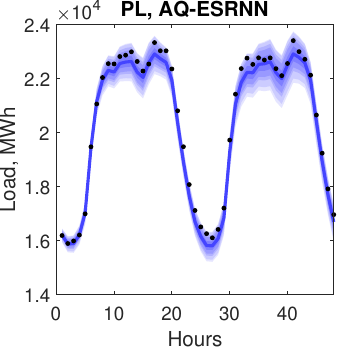}
    \includegraphics[width=0.248\linewidth]{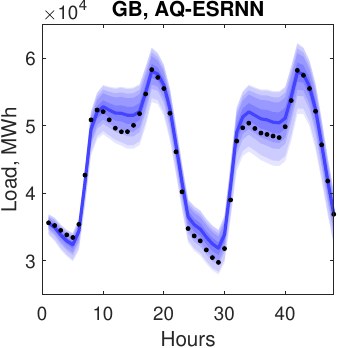}
    \includegraphics[width=0.24\linewidth]{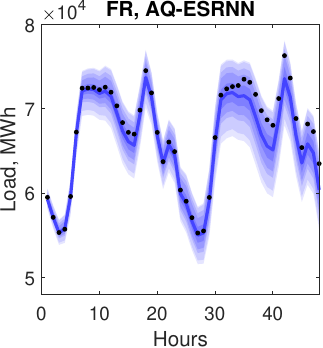}
    \includegraphics[width=0.24\linewidth]{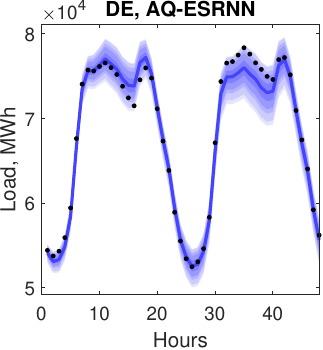}
    \includegraphics[width=0.248\linewidth]{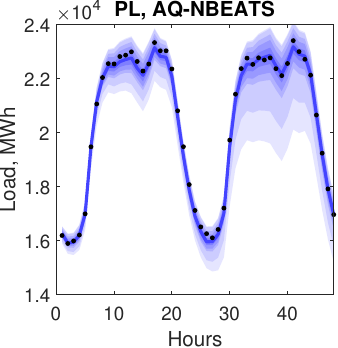}
    \includegraphics[width=0.248\linewidth]{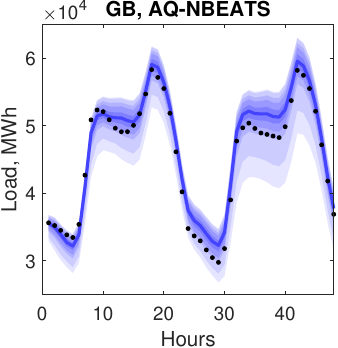}
    \includegraphics[width=0.24\linewidth]{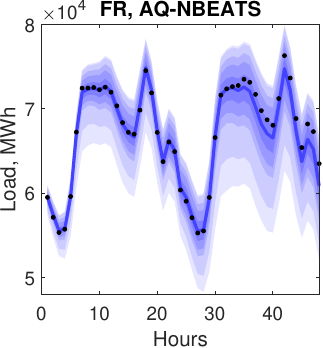}
    \includegraphics[width=0.24\linewidth]{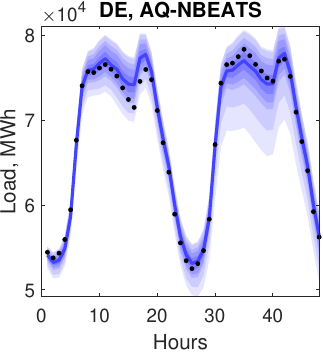}
    \caption{Examples of quantile forecasts for 30 and 31 January 2018 (50\%, 80\%, 90\%, 98\% and 99.8\% PIs are shown; dots represent true values).}
    \label{fig:For}
\end{figure}

To provide for a more comprehensive assessment of probabilistic forecasts, we additionally examine relative frequency (RF), also referred to as calibration or coverage~\cite{Mar22}. $RF(q)$ quantifies the proportion of observed values $y_i$ falling below or equal to the predicted $q$-quantile $\widehat{y}_{q,i}$ and is defined as follows: 
\begin{equation}
RF(q)=\frac{1}{M} \sum_{i=1}^M \mathds{1}{\{y_i \leq 
\widehat{y}_{q,i}\}}\,.
\label{eqrf}
\end{equation}
For probabilistic forecasts to be precise and informative, they should exhibit an RF value close to the nominal probability level, with a reasonably small deviation from the realized values~\cite{Gne07}. In essence, probabilistic forecasts are expected to capture uncertainty without significantly deviating from the actual series values. 
Ideally, predicted $q$-quantiles should surpass the realized values in $100 \cdot q\,\%$ of cases, ensuring an RF of $q$. To evaluate the average deviation of $RF(q)$ from the desired $q$ across all assumed quantile probabilities $q \in \Pi=\{0.001, 0.01, 0.02, ..., 0.99, 0.999\}$, we introduce the mean absolute RF error:
\begin{equation}
MARFE=\frac{1}{|\Pi|} \sum_{q \in \Pi} |RF(q)-q| \,.
\label{eqmrf}
\end{equation}
Figure \ref{fig:RF} illustrates the RF charts, where the desired RF values are depicted with dashed lines. Each dot represents RF calculated for a specific country and probability $q \in \Pi$, while the red line signifies the mean RF value across all countries. From this figure, it is apparent that DeepAR, WaveNet, Transformer and TFT exhibit the most distorted RF distributions, displaying deviations from the desired probabilities in various directions. MARFE, corresponding to the deviation of the red line from the dashed line, exceeds 0.04 for these models. In contrast, Naive and statistical models demonstrate less scattered results, with a slightly s-shaped curve. The lowest MARFE is observed for AQ-ESRNN (0.0157), followed by MLP (0.0168), Theta (0.0187) and AQ-NBEATS (0.0211). 
Notably, our any-quantile models accurately express the target probabilities, with RF curves closely aligned with the desired values and the average line almost coinciding with the dashed line.

Figure \ref{fig:For} concludes our empirical study by providing the examples of probabilistic forecasts generated by our proposed AQ-models for Poland, Great Britain, France and Germany.

\begin{figure}[t]
    \centering
    \includegraphics[width=1\linewidth]{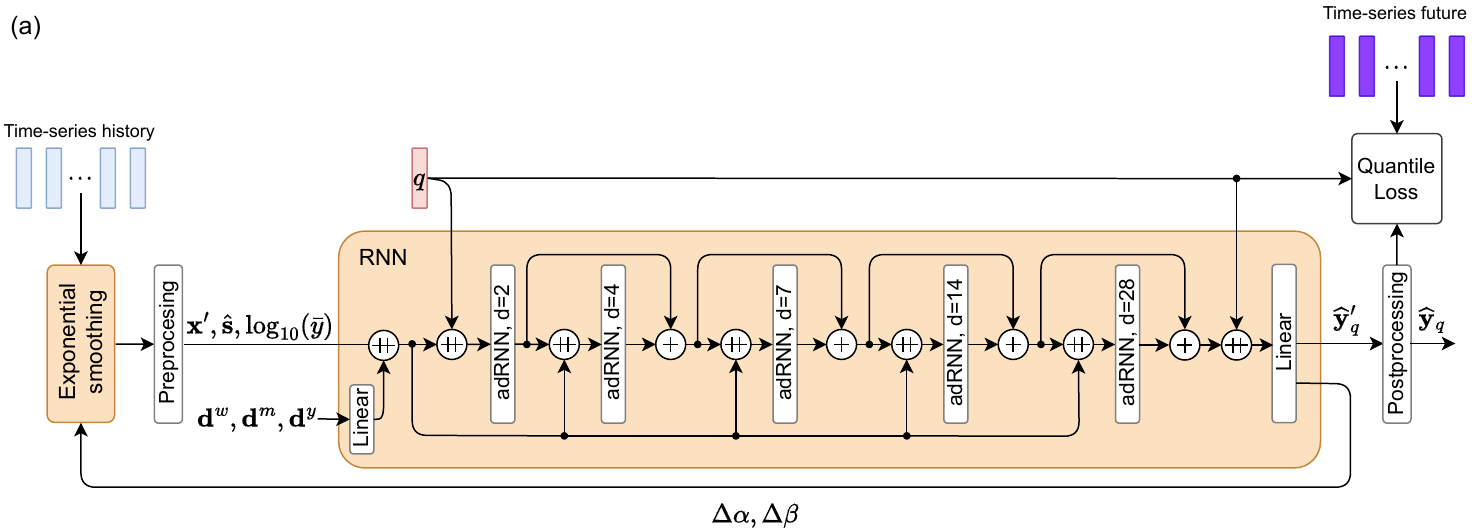}
    \includegraphics[width=1\linewidth]{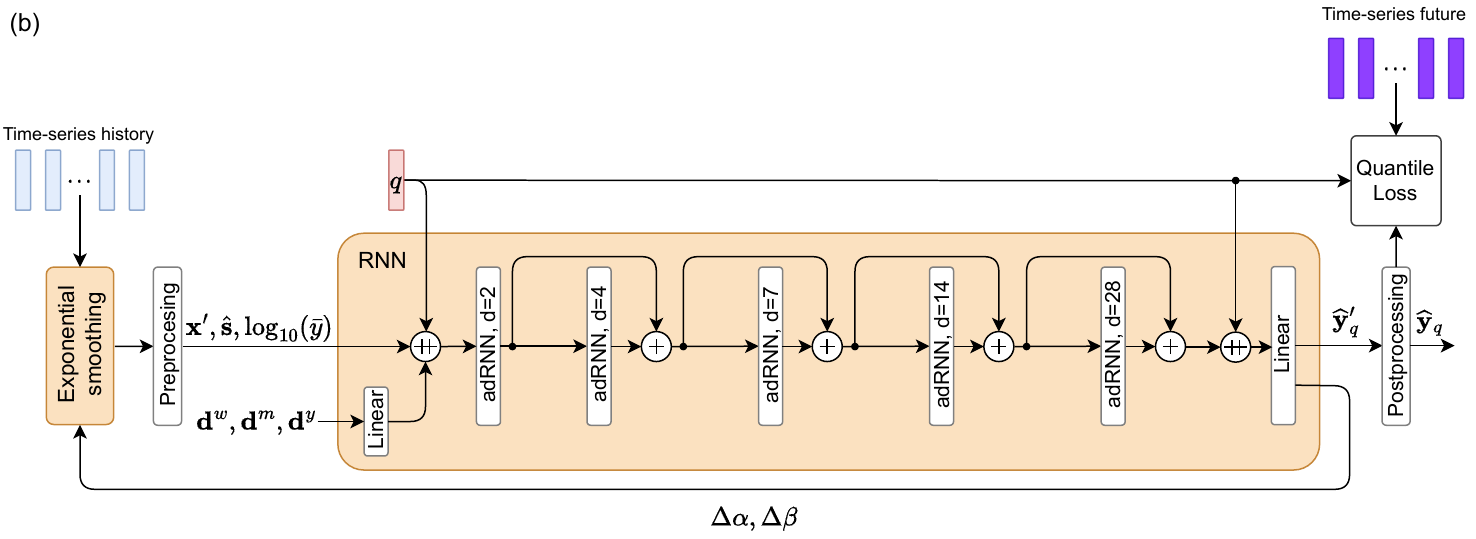} 
    \caption{Reduced versions of AQ-ESRNN: (a) AQ-ESRNN-noQuantInMidLayers, (b) AQ-ESRNN-noInputInMidLayers.}
    \label{fig:AQESRNN_abl}
\end{figure}

\subsection{AQ-ESRNN Ablation Studies}

\begin{table*}[t]
\centering
\caption{Ablation study of AQ-ESRNN.}
\begin{tabularx}{\textwidth}{l@{\extracolsep{\fill}} cccccc}
& CRPS & N-CRPS & MAPE  \\
\midrule
AQ-ESRNN & 195.9 & 1.72 & 2.32 \\
AQ-ESRNN, 4 Layers &  198.7 & 1.75 & 2.42 \\
AQ-ESRNN, 3 Layers &  202.7  & 1.77 & 2.41 \\
AQ-ESRNN-noQuantInMidLayers &  203.7  & 1.78 & 2.37 \\
AQ-ESRNN-noInputInMidLayers &  204.2  & 1.79 & 2.41 \\
AQ-ESRNN-dRNN &  199.2  & 1.74 & 2.36 \\
AQ-ESRNN-LSTM &  204.0 & 1.79  & 2.41 \\
AQ-ESRNN-unsorted & 264.2 & 2.34 & 2.35 \\
\bottomrule
\end{tabularx}%
\label{table:esrnn_ablation}
\end{table*}%

Ablation studies of AQ-ESRNN are reported in Table~\ref{table:esrnn_ablation}, with the full model showcased in the first row. Initially, we explore the impact of reducing the number of layers (cells). It is evident that a larger number of layers enhances accuracy, albeit at the expense of increased computation time. Next, AQ-ESRNN-noQuantInMidLayers removes the quantile from the middle layers, while AQ-ESRNN-noInputInMidLayers eliminates all inputs from the middle layers. These architectural variants are depicted in Figures~\ref{fig:AQESRNN_abl}a and~\ref{fig:AQESRNN_abl}b, respectively. Both of these simplified variants yield inferior results, demonstrating the value of ``reminding'' the target probability to the neural network at each block. Furthermore, AQ-ESRNN-dRNN variant represents the full model (5 layers) but employs simpler dRNN cells without the attention mechanism, as described in ~\cite{smyl2022dyn}. AQ-ESRNN-LSTM uses the standard LSTM cell. Simplifying the cell type leads to a deterioration in results. Lastly, AQ-ESRNN-unsorted is the full AQ-ESRNN model that does not sort forecast values. Sorting emerges as a crucial element of this model's CRPS performance. Without sorting, CRPS increases significantly, by approximately 34\%, compared to the full variant.

\subsection{AQ-NBEATS Ablation Studies}

Table~\ref{table:nbeats_ablation} studies the influence of the following factors in AQ-NBEATS architecture. First, AQ-NBEATS can work with both normalized and non-normalized inputs. Therefore, we explore the case of normalized input, in which case the input time-series is divided by the absolute maximum value of the input window (Max-Norm column checkmark) and unnormalized input  (Max-Norm column no checkmark). Second, learning supervision can be done either with the MQ loss defined in equation~\eqref{eqn:quantile_loss} (no checkmark in MQN column) or with its normalized MQN variant defined in equation~\eqref{eqn:mqn-loss-definition} (checkmark in MQN column). Finally, we explore the following AQ-NBEATS variants discussed in Section~\ref{ssec:aq_nbeats} and depicted in Figure~\ref{fig:any_quantile_variants}: (i) AQ-NBEATS-CAT, the quantile level is appended to the input sequence as additional token, (ii) AQ-NBEATS-FiLM, the quantile is inserted in every block through FiLM mechanism, (iii) AQ-NBEATS-OUT, a separate quantile block is appended at the end of the standard N-BEATS.

\begin{table*}[t]
\centering
\caption{Ablation study of the Max-Norm and the concatenation/FiLM based quantile incorporation mechanisms in the N-BEATS architecture.}
\begin{tabularx}{\textwidth}{l@{\extracolsep{\fill}} cccccc}
& Max-Norm & MQN & CRPS & N-CRPS & MAPE  \\
\midrule
AQ-NBEATS-CAT & \checkmark & & 215.3 & 1.94 & 2.62 \\
AQ-NBEATS-CAT &  & & 286.7 & 2.58 & 2.57 \\
AQ-NBEATS-FiLM & \checkmark  & & 219.3 & 1.98 & 2.68 \\
AQ-NBEATS-FiLM &  & & 214.1 & 1.86 & 2.51 \\
AQ-NBEATS-OUT & \checkmark  & & 215.7 & 1.94 & 2.61 \\
AQ-NBEATS-OUT &  & & 217.2 & 1.9 & 2.56 \\
\midrule
AQ-NBEATS-CAT & \checkmark & \checkmark & 215.5 & 1.86 & 2.52 \\
AQ-NBEATS-CAT &  & \checkmark & 285.1 & 2.46 & 2.48 \\
AQ-NBEATS-FiLM & \checkmark  & \checkmark & 228.1 & 1.94 & 2.62 \\
AQ-NBEATS-FiLM &  & \checkmark & 211.2 & 1.84 & 2.47  \\
AQ-NBEATS-OUT & \checkmark  & \checkmark & 216.4 & 1.86 & 2.52 \\
AQ-NBEATS-OUT &  & \checkmark & 212.5 & 1.84 & 2.48 \\
\bottomrule
\end{tabularx}%
\label{table:nbeats_ablation}
\end{table*}%

Looking at Table~\ref{table:nbeats_ablation} the following conclusions can be made. First, AQ-NBEATS-FiLM is the most accurate variant, which implies that providing quantile level as input to every block of the architecture is beneficial from the accuracy standpoint. Second, AQ-NBEATS-FiLM and AQ-NBEATS-OUT variants are significantly less sensitive to input normalization (Max-Norm) than the AQ-NBEATS-CAT variant. The main reason for this is that the time-series in the dataset have large magnitude. The quantile probability, distributed in $[0,1]$ range, when concatenated with a vector of large magnitude values, has a tendency to be neglected. Based on these observations, we believe that AQ-NBEATS-FiLM and AQ-NBEATS-OUT variants should be preferred in architecture designs due to robustness, accuracy and computational efficiency considerations. Indeed, recall that AQ-NBEATS-CAT deals with the $B \times T \times Q$ (batch, time, quantiles) input tensor due to the concatenation of quantiles, effectively multiplying the inference compute cost by a factor of $Q$. AQ-NBEATS-FiLM has similar compute explosion, but clearly better accuracy. AQ-NBEATS-OUT has comparable accuracy, but compute cost is multiplied by a factor of $Q$ only for one of the blocks, making it order of magnitude more compute efficient. Finally, the normalized MQ loss defined in~\eqref{eqn:mqn-loss-definition} may provide improvements, depending on architecture configurations, and can be considered as a hyperparameter to obtain incremental improvements in accuracy.

\section{Discussion of Findings} \label{Discussion}

\textbf{Flexibility of AQ-approach to Probabilistic Forecasting}. The review of related work on probabilistic forecasting highlights diverse approaches in Section~\ref{ssec:related_work}. These can be broadly categorized into (i) methods focusing on non-parametric quantile regression with fixed quantile probabilities and (ii) methods that adopt a parametric distribution models allowing them to predict the full distribution and thus arbitrary quantiles. Our methodology falls within the realm of non-parametric models. As such, it can model arbitrary conditional distributions. Yet, it retains the flexibility inherent to parametric techniques by being able to predict an arbitrary quantile. This is achieved through (i) the general parameterization of the forecaster $f_{\theta}( \vec{x}, q)$, treating the quantile probability $q$ as an input parameter and (ii) the learning methodology that trains the model to forecast future trajectories based on input patterns $\vec{x}$ conditioned on randomly selected quantile probabilities $q$. Consequently, our AQ-models offer enhanced flexibility and accuracy compared to classical and deep probabilistic forecasting baselines, which is supported by our experimental results.

\textbf{Quantile Injection Mechanism}. We experimented with different mechanisms used to inject the quantile probability value in the network. Results for both ESRNN and N-BEATS architectures seem to indicate that the accuracy-optimal injection mechanism exposes every block of the deep neural network to the value of desired quantile probability. The application of quantile only at the input and output works for both ESRNN and N-BEATS, being marginally worse. Additionally, both simple concatenation and more sophisticated mechanisms such as FiLM layer seem viable for injecting quantile probability. At the same time, concatenation works well in the case when input signal normalization is applied (which is the case of ESRNN by design and it can be used for N-BEATS additionally). FiLM works fine with both normalized and unnormalized data. Finally, injecting quantile probability at the last block of a deep learning architecture seems to be most computationally effective. Given that this is only marginally worse than injecting the quantile at every intermediate architectural block, this method can be preferable in applications requiring fast inference speed. We conjecture that this result implies that most of time-series representation work is independent of the quantile that the network targets. Therefore, the task of time-series representation and the task of quantile generation are largely disjoint and can be handled in a substantially independent way within the forecasting backbone, leading to inference speed gains.

\textbf{Quantile Crossing and Sorting}. In general, machine learning models trained for quantile regression may exhibit undesirable quantile crossing behavior, where a forecast for a lower quantile exceeds that of a higher quantile. While the frequency of these occurrences can be reduced by augmenting the loss function with a suitable penalty \cite{tang2022}, this does not guarantee that quantile crossing will never occur. Achieving this requires a more complex model, as demonstrated in studies such as \cite{park2022} and \cite{tang2022}. Alternatively, post-processing methods can be applied, such as sorting or isotonic projection \cite{fakoor2023flexible}. Sorting is the simplest method, but it has been shown to have desirable properties, such as ensuring that CRPS never decreases \cite{chernozhukov2010}.
Interestingly, the two models proposed in this paper, AQ-ESRNN and AQ-NBEATS, respond differently to sorting: it is a crucial step for the former (the metrics improve by 30\%, see Table~\ref{table:esrnn_ablation}) and is practically unimportant for the latter. We leave the in-depth study of root-causes of this phenomenon for future work. In the current work, we conclude that, generally speaking, sorting is a powerful mechanism for improving multi-quantile forecasting, whereas its effectiveness may vary depending on underlying model and problem.

\textbf{Probability Sampling Distribution}. Another distinction between the two models lies in the sampling distribution utilized for generating probabilities during training. AQ-ESRNN employs Beta(0.3, 0.3) (please refer to Fig.~\ref{fig:beta}), while AQ-NBEATS remains unaffected by this innovation and utilizes the uniform distribution. Originally, the idea of employing a convex Beta distribution for AQ-ESRNN was introduced to enhance the accuracy of extreme quantiles (where probability is close to 0 or 1), with the expectation that improvements in accuracy in these areas would negatively impact overall accuracy. However, it transpired that this approach also benefits the overall distribution metric, such as CRPS. We speculate that this difference between the two models stems from the fact that AQ-NBEATS employs ReLU nonlinearity, which allows the output to easily grow to any required value, while AQ-ESRNN employs sigmoid and tanh nonlinearities, which may pose more challenges for the output to reach extreme values.

\section{Conclusion} \label{Con}

We present a general probabilistic forecasting methodology applicable to a large class of neural network forecasting algorithms. At training time, the methodology exposes a learnable forecasting algorithm to the task of predicting a randomly sampled quantile. When trained on this task, forecasting algorithm learns to provide an arbitrary quantile at inference time. We apply this general methodology to two well-known neural forecasting algorithms via novel architectural modifications. We empirically show that the proposed modified algorithms perform substantially better than a number of classical and neural forecasting baselines. Moving forward, our research will focus on extending our current results to additional neural network architectures, other machine learning models as well as showing the effectiveness of the proposed methodology across different application domains.

\bibliographystyle{elsarticle-num-names}
\bibliography{main}

\begin{thebibliography}{62}
\expandafter\ifx\csname natexlab\endcsname\relax\def\natexlab#1{#1}\fi
\providecommand{\url}[1]{\texttt{#1}}
\providecommand{\href}[2]{#2}
\providecommand{\path}[1]{#1}
\providecommand{\DOIprefix}{doi:}
\providecommand{\ArXivprefix}{arXiv:}
\providecommand{\URLprefix}{URL: }
\providecommand{\Pubmedprefix}{pmid:}
\providecommand{\doi}[1]{\href{http://dx.doi.org/#1}{\path{#1}}}
\providecommand{\Pubmed}[1]{\href{pmid:#1}{\path{#1}}}
\providecommand{\bibinfo}[2]{#2}
\ifx\xfnm\relax \def\xfnm[#1]{\unskip,\space#1}\fi
%Type = Article
\bibitem[{Haupt et~al.(2019)Haupt, Garcia~Casado, Davidson, Dobschinski, Du,
  Lange, Miller, Mohrlen, Motley, Pestana, and Zack}]{Hau19}
\bibinfo{author}{S.~E. Haupt}, \bibinfo{author}{M.~Garcia~Casado},
  \bibinfo{author}{M.~Davidson}, \bibinfo{author}{J.~Dobschinski},
  \bibinfo{author}{P.~Du}, \bibinfo{author}{M.~Lange},
  \bibinfo{author}{T.~Miller}, \bibinfo{author}{C.~Mohrlen},
  \bibinfo{author}{A.~Motley}, \bibinfo{author}{R.~Pestana},
  \bibinfo{author}{J.~Zack},
\newblock \bibinfo{title}{The use of probabilistic forecasts: Applying them in
  theory and practice},
\newblock \bibinfo{journal}{IEEE Power and Energy Magazine}
  \bibinfo{volume}{17} (\bibinfo{year}{2019}) \bibinfo{pages}{46--57}.
  \DOIprefix\doi{10.1109/MPE.2019.2932639}.
%Type = Inproceedings
\bibitem[{Etingov et~al.(2018)Etingov, Miller, Hou, Makarov, Pennock, Beaucage,
  Loutan, and Motley}]{Eti18}
\bibinfo{author}{P.~Etingov}, \bibinfo{author}{L.~Miller},
  \bibinfo{author}{Z.~Hou}, \bibinfo{author}{Y.~Makarov},
  \bibinfo{author}{K.~Pennock}, \bibinfo{author}{P.~Beaucage},
  \bibinfo{author}{C.~Loutan}, \bibinfo{author}{A.~Motley},
\newblock \bibinfo{title}{Balancing needs assessment using advanced
  probabilistic forecasts},
\newblock in: \bibinfo{booktitle}{2018 IEEE International Conference on
  Probabilistic Methods Applied to Power Systems (PMAPS)},
  \bibinfo{year}{2018}, pp. \bibinfo{pages}{1--6}.
  \DOIprefix\doi{10.1109/PMAPS.2018.8440392}.
%Type = Article
\bibitem[{Wang et~al.(2023)Wang, Tuohy, Ortega-Vazquez, Bello, Ela,
  Kirk-Davidoff, Hobbs, Ault, and Philbrick}]{Wan23}
\bibinfo{author}{Q.~Wang}, \bibinfo{author}{A.~Tuohy},
  \bibinfo{author}{M.~Ortega-Vazquez}, \bibinfo{author}{M.~Bello},
  \bibinfo{author}{E.~Ela}, \bibinfo{author}{D.~Kirk-Davidoff},
  \bibinfo{author}{W.~B. Hobbs}, \bibinfo{author}{D.~J. Ault},
  \bibinfo{author}{R.~Philbrick},
\newblock \bibinfo{title}{Quantifying the value of probabilistic forecasting
  for power system operation planning},
\newblock \bibinfo{journal}{Applied Energy} \bibinfo{volume}{343}
  (\bibinfo{year}{2023}) \bibinfo{pages}{121254}. \URLprefix
  \url{https://www.sciencedirect.com/science/article/pii/S0306261923006189}.
  \DOIprefix\doi{10.1016/j.apenergy.2023.121254}.
%Type = Article
\bibitem[{Yamujala et~al.(2022)Yamujala, Jain, Sreekumar, Bhakar, and
  Mathur}]{Yam22}
\bibinfo{author}{S.~Yamujala}, \bibinfo{author}{A.~Jain},
  \bibinfo{author}{S.~Sreekumar}, \bibinfo{author}{R.~Bhakar},
  \bibinfo{author}{J.~Mathur},
\newblock \bibinfo{title}{Enhancing power systems operational flexibility with
  ramp products from flexible resources},
\newblock \bibinfo{journal}{Electric Power Systems Research}
  \bibinfo{volume}{202} (\bibinfo{year}{2022}) \bibinfo{pages}{107599}.
  \URLprefix
  \url{https://www.sciencedirect.com/science/article/pii/S0378779621005800}.
  \DOIprefix\doi{10.1016/j.epsr.2021.107599}.
%Type = Article
\bibitem[{Le et~al.(2016)Le, Berizzi, and Bovo}]{Le16}
\bibinfo{author}{D.~Le}, \bibinfo{author}{A.~Berizzi},
  \bibinfo{author}{C.~Bovo},
\newblock \bibinfo{title}{A probabilistic security assessment approach to power
  systems with integrated wind resources},
\newblock \bibinfo{journal}{Renewable Energy} \bibinfo{volume}{85}
  (\bibinfo{year}{2016}) \bibinfo{pages}{114--123}. \URLprefix
  \url{https://www.sciencedirect.com/science/article/pii/S0960148115300598}.
  \DOIprefix\doi{10.1016/j.renene.2015.06.035}.
%Type = Inproceedings
\bibitem[{Beykirch et~al.(2022)Beykirch, Janke, and Steinke}]{Bey22}
\bibinfo{author}{M.~Beykirch}, \bibinfo{author}{T.~Janke},
  \bibinfo{author}{F.~Steinke},
\newblock \bibinfo{title}{Bidding and scheduling in energy markets: Which
  probabilistic forecast do we need?},
\newblock in: \bibinfo{booktitle}{2022 17th International Conference on
  Probabilistic Methods Applied to Power Systems (PMAPS)},
  \bibinfo{year}{2022}, pp. \bibinfo{pages}{1--6}.
  \DOIprefix\doi{10.1109/PMAPS53380.2022.9810632}.
%Type = Article
\bibitem[{Hong and Fan(2016)}]{Hon16}
\bibinfo{author}{T.~Hong}, \bibinfo{author}{S.~Fan},
\newblock \bibinfo{title}{Probabilistic electric load forecasting: A tutorial
  review},
\newblock \bibinfo{journal}{International Journal of Forecasting}
  \bibinfo{volume}{32} (\bibinfo{year}{2016}) \bibinfo{pages}{914--938}.
  \URLprefix
  \url{https://www.sciencedirect.com/science/article/pii/S0169207015001508}.
  \DOIprefix\doi{10.1016/j.ijforecast.2015.11.011}.
%Type = Article
\bibitem[{Khajeh and Laaksonen(2022)}]{Kha22}
\bibinfo{author}{H.~Khajeh}, \bibinfo{author}{H.~Laaksonen},
\newblock \bibinfo{title}{Applications of probabilistic forecasting in smart
  grids: A review},
\newblock \bibinfo{journal}{Applied Sciences} \bibinfo{volume}{12}
  (\bibinfo{year}{2022}). \URLprefix
  \url{https://www.mdpi.com/2076-3417/12/4/1823}.
  \DOIprefix\doi{10.3390/app12041823}.
%Type = Article
\bibitem[{Li and Zhang(2020)}]{Li20}
\bibinfo{author}{B.~Li}, \bibinfo{author}{J.~Zhang},
\newblock \bibinfo{title}{A review on the integration of probabilistic solar
  forecasting in power systems},
\newblock \bibinfo{journal}{Solar Energy} \bibinfo{volume}{210}
  (\bibinfo{year}{2020}) \bibinfo{pages}{68--86}. \URLprefix
  \url{https://www.sciencedirect.com/science/article/pii/S0038092X20307982}.
  \DOIprefix\doi{10.1016/j.solener.2020.07.066}, \bibinfo{note}{special Issue
  on Grid Integration}.
%Type = Article
\bibitem[{Hyndman and Fan(2010)}]{Hyn10}
\bibinfo{author}{R.~J. Hyndman}, \bibinfo{author}{S.~Fan},
\newblock \bibinfo{title}{Density forecasting for long-term peak electricity
  demand},
\newblock \bibinfo{journal}{IEEE Transactions on Power Systems}
  \bibinfo{volume}{25} (\bibinfo{year}{2010}) \bibinfo{pages}{1142--1153}.
  \DOIprefix\doi{10.1109/TPWRS.2009.2036017}.
%Type = Article
\bibitem[{Taylor and Buizza(2002)}]{Tay02}
\bibinfo{author}{J.~W. Taylor}, \bibinfo{author}{R.~Buizza},
\newblock \bibinfo{title}{Neural network load forecasting with weather ensemble
  predictions},
\newblock \bibinfo{journal}{IEEE Transactions on Power Systems}
  \bibinfo{volume}{17} (\bibinfo{year}{2002}) \bibinfo{pages}{626 – 632}.
  \URLprefix
  \url{https://www.scopus.com/inward/record.uri?eid=2-s2.0-0036699869&doi=10.1109%2fTPWRS.2002.800906&partnerID=40&md5=019f35fba591b6a47c11ac1e27c6fdbe}.
  \DOIprefix\doi{10.1109/TPWRS.2002.800906}.
%Type = Article
\bibitem[{Khoshrou and Pauwels(2019)}]{Kho19}
\bibinfo{author}{A.~Khoshrou}, \bibinfo{author}{E.~J. Pauwels},
\newblock \bibinfo{title}{Short-term scenario-based probabilistic load
  forecasting: A data-driven approach},
\newblock \bibinfo{journal}{Applied Energy} \bibinfo{volume}{238}
  (\bibinfo{year}{2019}) \bibinfo{pages}{1258--1268}. \URLprefix
  \url{https://www.sciencedirect.com/science/article/pii/S0306261919301412}.
  \DOIprefix\doi{10.1016/j.apenergy.2019.01.155}.
%Type = Article
\bibitem[{Xie and Hong(2018)}]{Xie18}
\bibinfo{author}{J.~Xie}, \bibinfo{author}{T.~Hong},
\newblock \bibinfo{title}{Temperature scenario generation for probabilistic
  load forecasting},
\newblock \bibinfo{journal}{IEEE Transactions on Smart Grid}
  \bibinfo{volume}{9} (\bibinfo{year}{2018}) \bibinfo{pages}{1680--1687}.
  \DOIprefix\doi{10.1109/TSG.2016.2597178}.
%Type = Article
\bibitem[{Xie et~al.(2017)Xie, Hong, Laing, and Kang}]{Xie17}
\bibinfo{author}{J.~Xie}, \bibinfo{author}{T.~Hong},
  \bibinfo{author}{T.~Laing}, \bibinfo{author}{C.~Kang},
\newblock \bibinfo{title}{On normality assumption in residual simulation for
  probabilistic load forecasting},
\newblock \bibinfo{journal}{IEEE Transactions on Smart Grid}
  \bibinfo{volume}{8} (\bibinfo{year}{2017}) \bibinfo{pages}{1046--1053}.
  \DOIprefix\doi{10.1109/TSG.2015.2447007}.
%Type = Inproceedings
\bibitem[{Hor et~al.(2006)Hor, Watson, and Majithia}]{Hor06}
\bibinfo{author}{C.-L. Hor}, \bibinfo{author}{S.~J. Watson},
  \bibinfo{author}{S.~Majithia},
\newblock \bibinfo{title}{Daily load forecasting and maximum demand estimation
  using arima and garch},
\newblock in: \bibinfo{booktitle}{2006 International Conference on
  Probabilistic Methods Applied to Power Systems}, \bibinfo{year}{2006}, pp.
  \bibinfo{pages}{1--6}. \DOIprefix\doi{10.1109/PMAPS.2006.360237}.
%Type = Article
\bibitem[{Cao et~al.(2020)Cao, Wan, Zhang, Li, and Song}]{Cao20}
\bibinfo{author}{Z.~Cao}, \bibinfo{author}{C.~Wan}, \bibinfo{author}{Z.~Zhang},
  \bibinfo{author}{F.~Li}, \bibinfo{author}{Y.~Song},
\newblock \bibinfo{title}{Hybrid ensemble deep learning for deterministic and
  probabilistic low-voltage load forecasting},
\newblock \bibinfo{journal}{IEEE Transactions on Power Systems}
  \bibinfo{volume}{35} (\bibinfo{year}{2020}) \bibinfo{pages}{1881--1897}.
  \DOIprefix\doi{10.1109/TPWRS.2019.2946701}.
%Type = Article
\bibitem[{Liu et~al.(2017)Liu, Nowotarski, Hong, and Weron}]{Liu17}
\bibinfo{author}{B.~Liu}, \bibinfo{author}{J.~Nowotarski},
  \bibinfo{author}{T.~Hong}, \bibinfo{author}{R.~Weron},
\newblock \bibinfo{title}{Probabilistic load forecasting via quantile
  regression averaging on sister forecasts},
\newblock \bibinfo{journal}{IEEE Transactions on Smart Grid}
  \bibinfo{volume}{8} (\bibinfo{year}{2017}) \bibinfo{pages}{730--737}.
  \DOIprefix\doi{10.1109/TSG.2015.2437877}.
%Type = Article
\bibitem[{Yang et~al.(2018)Yang, Li, Li, and Qu}]{Yan18}
\bibinfo{author}{Y.~Yang}, \bibinfo{author}{S.~Li}, \bibinfo{author}{W.~Li},
  \bibinfo{author}{M.~Qu},
\newblock \bibinfo{title}{Power load probability density forecasting using
  gaussian process quantile regression},
\newblock \bibinfo{journal}{Applied Energy} \bibinfo{volume}{213}
  (\bibinfo{year}{2018}) \bibinfo{pages}{499--509}. \URLprefix
  \url{https://www.sciencedirect.com/science/article/pii/S0306261917316100}.
  \DOIprefix\doi{10.1016/j.apenergy.2017.11.035}.
%Type = Article
\bibitem[{He et~al.(2019)He, Qin, Wang, Wang, and Wang}]{He19}
\bibinfo{author}{Y.~He}, \bibinfo{author}{Y.~Qin}, \bibinfo{author}{S.~Wang},
  \bibinfo{author}{X.~Wang}, \bibinfo{author}{C.~Wang},
\newblock \bibinfo{title}{Electricity consumption probability density
  forecasting method based on lasso-quantile regression neural network},
\newblock \bibinfo{journal}{Applied Energy} \bibinfo{volume}{233-234}
  (\bibinfo{year}{2019}) \bibinfo{pages}{565--575}. \URLprefix
  \url{https://www.sciencedirect.com/science/article/pii/S0306261918316301}.
  \DOIprefix\doi{10.1016/j.apenergy.2018.10.061}.
%Type = Article
\bibitem[{Liu et~al.(2022)Liu, Chen, Sun, Muyeen, Lin, and Mi}]{Liu22}
\bibinfo{author}{R.~Liu}, \bibinfo{author}{T.~Chen}, \bibinfo{author}{G.~Sun},
  \bibinfo{author}{S.~Muyeen}, \bibinfo{author}{S.~Lin},
  \bibinfo{author}{Y.~Mi},
\newblock \bibinfo{title}{Short-term probabilistic building load forecasting
  based on feature integrated artificial intelligent approach},
\newblock \bibinfo{journal}{Electric Power Systems Research}
  \bibinfo{volume}{206} (\bibinfo{year}{2022}) \bibinfo{pages}{107802}.
  \URLprefix
  \url{https://www.sciencedirect.com/science/article/pii/S0378779622000323}.
  \DOIprefix\doi{10.1016/j.epsr.2022.107802}.
%Type = Article
\bibitem[{He et~al.(2022)He, Cao, Wang, and Fu}]{He22}
\bibinfo{author}{Y.~He}, \bibinfo{author}{C.~Cao}, \bibinfo{author}{S.~Wang},
  \bibinfo{author}{H.~Fu},
\newblock \bibinfo{title}{Nonparametric probabilistic load forecasting based on
  quantile combination in electrical power systems},
\newblock \bibinfo{journal}{Applied Energy} \bibinfo{volume}{322}
  (\bibinfo{year}{2022}) \bibinfo{pages}{119507}. \URLprefix
  \url{https://www.sciencedirect.com/science/article/pii/S0306261922008273}.
  \DOIprefix\doi{10.1016/j.apenergy.2022.119507}.
%Type = Article
\bibitem[{Li et~al.(2023)Li, Mo, Gao, and Bai}]{Li23}
\bibinfo{author}{B.~Li}, \bibinfo{author}{Y.~Mo}, \bibinfo{author}{F.~Gao},
  \bibinfo{author}{X.~Bai},
\newblock \bibinfo{title}{Short-term probabilistic load forecasting method
  based on uncertainty estimation and deep learning model considering
  meteorological factors},
\newblock \bibinfo{journal}{Electric Power Systems Research}
  \bibinfo{volume}{225} (\bibinfo{year}{2023}) \bibinfo{pages}{109804}.
  \URLprefix
  \url{https://www.sciencedirect.com/science/article/pii/S0378779623006934}.
  \DOIprefix\doi{10.1016/j.epsr.2023.109804}.
%Type = Article
\bibitem[{He et~al.(2017)He, Liu, Li, Wang, and Lu}]{He17}
\bibinfo{author}{Y.~He}, \bibinfo{author}{R.~Liu}, \bibinfo{author}{H.~Li},
  \bibinfo{author}{S.~Wang}, \bibinfo{author}{X.~Lu},
\newblock \bibinfo{title}{Short-term power load probability density forecasting
  method using kernel-based support vector quantile regression and copula
  theory},
\newblock \bibinfo{journal}{Applied Energy} \bibinfo{volume}{185}
  (\bibinfo{year}{2017}) \bibinfo{pages}{254--266}. \URLprefix
  \url{https://www.sciencedirect.com/science/article/pii/S0306261916315239}.
  \DOIprefix\doi{10.1016/j.apenergy.2016.10.079}.
%Type = Article
\bibitem[{He et~al.(2022)He, Xiao, An, Cao, and Xiao}]{He22a}
\bibinfo{author}{Y.~He}, \bibinfo{author}{J.~Xiao}, \bibinfo{author}{X.~An},
  \bibinfo{author}{C.~Cao}, \bibinfo{author}{J.~Xiao},
\newblock \bibinfo{title}{Short-term power load probability density forecasting
  based on {GLRQ}-stacking ensemble learning method},
\newblock \bibinfo{journal}{International Journal of Electrical Power \& Energy
  Systems} \bibinfo{volume}{142} (\bibinfo{year}{2022})
  \bibinfo{pages}{108243}. \URLprefix
  \url{https://www.sciencedirect.com/science/article/pii/S0142061522002721}.
  \DOIprefix\doi{10.1016/j.ijepes.2022.108243}.
%Type = Inproceedings
\bibitem[{Dudek(2024)}]{Dud24}
\bibinfo{author}{G.~Dudek},
\newblock \bibinfo{title}{Stacking for probabilistic short-term load
  forecasting},
\newblock in: \bibinfo{booktitle}{Int. Conf. on Computational Science, ICCS
  2024 (in print)}, \bibinfo{year}{2024}.
%Type = Article
\bibitem[{Hong et~al.(2020)Hong, Pinson, Wang, Weron, Yang, and
  Zareipour}]{Hon20}
\bibinfo{author}{T.~Hong}, \bibinfo{author}{P.~Pinson},
  \bibinfo{author}{Y.~Wang}, \bibinfo{author}{R.~Weron},
  \bibinfo{author}{D.~Yang}, \bibinfo{author}{H.~Zareipour},
\newblock \bibinfo{title}{Energy forecasting: A review and outlook},
\newblock \bibinfo{journal}{IEEE Open Access Journal of Power and Energy}
  \bibinfo{volume}{7} (\bibinfo{year}{2020}) \bibinfo{pages}{376--388}.
  \DOIprefix\doi{10.1109/OAJPE.2020.3029979}.
%Type = Article
\bibitem[{Wang et~al.(2019)Wang, Gan, Sun, Zhang, Lu, and Kang}]{Wan19}
\bibinfo{author}{Y.~Wang}, \bibinfo{author}{D.~Gan}, \bibinfo{author}{M.~Sun},
  \bibinfo{author}{N.~Zhang}, \bibinfo{author}{Z.~Lu},
  \bibinfo{author}{C.~Kang},
\newblock \bibinfo{title}{Probabilistic individual load forecasting using
  pinball loss guided lstm},
\newblock \bibinfo{journal}{Applied Energy} \bibinfo{volume}{235}
  (\bibinfo{year}{2019}) \bibinfo{pages}{10--20}. \URLprefix
  \url{https://www.sciencedirect.com/science/article/pii/S0306261918316465}.
  \DOIprefix\doi{10.1016/j.apenergy.2018.10.078}.
%Type = Article
\bibitem[{Zhang et~al.(2019)Zhang, Quan, and Srinivasan}]{Zha19}
\bibinfo{author}{W.~Zhang}, \bibinfo{author}{H.~Quan},
  \bibinfo{author}{D.~Srinivasan},
\newblock \bibinfo{title}{An improved quantile regression neural network for
  probabilistic load forecasting},
\newblock \bibinfo{journal}{IEEE Transactions on Smart Grid}
  \bibinfo{volume}{10} (\bibinfo{year}{2019}) \bibinfo{pages}{4425--4434}.
  \DOIprefix\doi{10.1109/TSG.2018.2859749}.
%Type = Article
\bibitem[{Li et~al.(2023)Li, Tan, Zhang, Miao, and He}]{Li23a}
\bibinfo{author}{D.~Li}, \bibinfo{author}{Y.~Tan}, \bibinfo{author}{Y.~Zhang},
  \bibinfo{author}{S.~Miao}, \bibinfo{author}{S.~He},
\newblock \bibinfo{title}{Probabilistic forecasting method for mid-term hourly
  load time series based on an improved temporal fusion transformer model},
\newblock \bibinfo{journal}{International Journal of Electrical Power \& Energy
  Systems} \bibinfo{volume}{146} (\bibinfo{year}{2023})
  \bibinfo{pages}{108743}. \URLprefix
  \url{https://www.sciencedirect.com/science/article/pii/S0142061522007396}.
  \DOIprefix\doi{10.1016/j.ijepes.2022.108743}.
%Type = Article
\bibitem[{Smyl et~al.(2023)Smyl, Dudek, and Pełka}]{Smy23}
\bibinfo{author}{S.~Smyl}, \bibinfo{author}{G.~Dudek},
  \bibinfo{author}{P.~Pełka},
\newblock \bibinfo{title}{Es-drnn: A hybrid exponential smoothing and dilated
  recurrent neural network model for short-term load forecasting},
\newblock \bibinfo{journal}{IEEE Transactions on Neural Networks and Learning
  Systems}  (\bibinfo{year}{2023}) \bibinfo{pages}{1--13}.
  \DOIprefix\doi{10.1109/TNNLS.2023.3259149}.
%Type = Article
\bibitem[{Alfieri and De~Falco(2020)}]{Alf20}
\bibinfo{author}{L.~Alfieri}, \bibinfo{author}{P.~De~Falco},
\newblock \bibinfo{title}{Wavelet-based decompositions in probabilistic load
  forecasting},
\newblock \bibinfo{journal}{IEEE Transactions on Smart Grid}
  \bibinfo{volume}{11} (\bibinfo{year}{2020}) \bibinfo{pages}{1367--1376}.
  \DOIprefix\doi{10.1109/TSG.2019.2937072}.
%Type = Article
\bibitem[{{van der Meer} et~al.(2018){van der Meer}, Shepero, Svensson, Widén,
  and Munkhammar}]{Mee18}
\bibinfo{author}{D.~{van der Meer}}, \bibinfo{author}{M.~Shepero},
  \bibinfo{author}{A.~Svensson}, \bibinfo{author}{J.~Widén},
  \bibinfo{author}{J.~Munkhammar},
\newblock \bibinfo{title}{Probabilistic forecasting of electricity consumption,
  photovoltaic power generation and net demand of an individual building using
  gaussian processes},
\newblock \bibinfo{journal}{Applied Energy} \bibinfo{volume}{213}
  (\bibinfo{year}{2018}) \bibinfo{pages}{195--207}. \URLprefix
  \url{https://www.sciencedirect.com/science/article/pii/S0306261917318275}.
  \DOIprefix\doi{10.1016/j.apenergy.2017.12.104}.
%Type = Article
\bibitem[{Alexandrov et~al.(2020)Alexandrov, Benidis, Bohlke-Schneider,
  Flunkert, Gasthaus, Januschowski, Maddix, Rangapuram, Salinas, and
  Schulz}]{Ale20}
\bibinfo{author}{A.~Alexandrov}, \bibinfo{author}{K.~Benidis},
  \bibinfo{author}{M.~Bohlke-Schneider}, \bibinfo{author}{V.~Flunkert},
  \bibinfo{author}{J.~Gasthaus}, \bibinfo{author}{T.~Januschowski},
  \bibinfo{author}{D.~C. Maddix}, \bibinfo{author}{S.~Rangapuram},
  \bibinfo{author}{D.~Salinas}, \bibinfo{author}{J.~Schulz},
\newblock \bibinfo{title}{Gluonts: Probabilistic and neural time series
  modeling in python},
\newblock \bibinfo{journal}{The Journal of Machine Learning Research}
  \bibinfo{volume}{21} (\bibinfo{year}{2020}) \bibinfo{pages}{4629–4634}.
%Type = Article
\bibitem[{Brusaferri et~al.(2022)Brusaferri, Matteucci, Spinelli, and
  Vitali}]{Bru22}
\bibinfo{author}{A.~Brusaferri}, \bibinfo{author}{M.~Matteucci},
  \bibinfo{author}{S.~Spinelli}, \bibinfo{author}{A.~Vitali},
\newblock \bibinfo{title}{Probabilistic electric load forecasting through
  bayesian mixture density networks},
\newblock \bibinfo{journal}{Applied Energy} \bibinfo{volume}{309}
  (\bibinfo{year}{2022}) \bibinfo{pages}{118341}. \URLprefix
  \url{https://www.sciencedirect.com/science/article/pii/S0306261921015907}.
  \DOIprefix\doi{10.1016/j.apenergy.2021.118341}.
%Type = Misc
\bibitem[{Gouttes et~al.(2021)Gouttes, Rasul, Koren, Stephan, and
  Naghibi}]{gouttes2021probabilistic}
\bibinfo{author}{A.~Gouttes}, \bibinfo{author}{K.~Rasul},
  \bibinfo{author}{M.~Koren}, \bibinfo{author}{J.~Stephan},
  \bibinfo{author}{T.~Naghibi}, \bibinfo{title}{Probabilistic time series
  forecasting with implicit quantile networks}, \bibinfo{year}{2021}.
  \href{http://arxiv.org/abs/2107.03743}{{\tt arXiv:2107.03743}}.
%Type = Inproceedings
\bibitem[{Karimi et~al.(2016)Karimi, Nutini, and Schmidt}]{Karimi2016linear}
\bibinfo{author}{H.~Karimi}, \bibinfo{author}{J.~Nutini},
  \bibinfo{author}{M.~Schmidt},
\newblock \bibinfo{title}{Linear convergence of gradient and proximal-gradient
  methods under the polyak-{\l}ojasiewicz condition},
\newblock in: \bibinfo{editor}{P.~Frasconi}, \bibinfo{editor}{N.~Landwehr},
  \bibinfo{editor}{G.~Manco}, \bibinfo{editor}{J.~Vreeken} (Eds.),
  \bibinfo{booktitle}{Machine Learning and Knowledge Discovery in Databases},
  \bibinfo{publisher}{Springer International Publishing},
  \bibinfo{address}{Cham}, \bibinfo{year}{2016}, pp. \bibinfo{pages}{795--811}.
  \DOIprefix\doi{10.1007/978-3-319-46128-1_50}.
%Type = Article
\bibitem[{Gneiting and Ranjan(2011)}]{tilmann2011comparing}
\bibinfo{author}{T.~Gneiting}, \bibinfo{author}{R.~Ranjan},
\newblock \bibinfo{title}{Comparing density forecasts using threshold-and
  quantile-weighted scoring rules},
\newblock \bibinfo{journal}{Journal of Business \& Economic Statistics}
  \bibinfo{volume}{29} (\bibinfo{year}{2011}) \bibinfo{pages}{411--422}.
  \DOIprefix\doi{10.1198/jbes.2010.08110}.
%Type = Inproceedings
\bibitem[{Smyl et~al.(2022)Smyl, Dudek, and Pelka}]{smyl2022dyn}
\bibinfo{author}{S.~Smyl}, \bibinfo{author}{G.~Dudek},
  \bibinfo{author}{P.~Pelka},
\newblock \bibinfo{title}{{ES-dRNN} with dynamic attention for short-term load
  forecasting},
\newblock in: \bibinfo{booktitle}{2022 International Joint Conference on Neural
  Networks (IJCNN)}, \bibinfo{year}{2022}, pp. \bibinfo{pages}{1--8}.
  \DOIprefix\doi{10.1109/IJCNN55064.2022.9889791}.
%Type = Inproceedings
\bibitem[{Oreshkin et~al.(2020)Oreshkin, Carpov, Chapados, and Bengio}]{Ore19}
\bibinfo{author}{B.~N. Oreshkin}, \bibinfo{author}{D.~Carpov},
  \bibinfo{author}{N.~Chapados}, \bibinfo{author}{Y.~Bengio},
\newblock \bibinfo{title}{{N-BEATS}: Neural basis expansion analysis for
  interpretable time series forecasting},
\newblock in: \bibinfo{booktitle}{ICLR}, \bibinfo{year}{2020}.
%Type = Book
\bibitem[{Hyndman and Athanasopoulos(2018)}]{Hyn20}
\bibinfo{author}{R.~Hyndman}, \bibinfo{author}{G.~Athanasopoulos},
  \bibinfo{title}{Forecasting: Principles and Practice}, \bibinfo{edition}{2nd}
  ed., \bibinfo{publisher}{OTexts}, \bibinfo{address}{Australia},
  \bibinfo{year}{2018}.
%Type = Article
\bibitem[{Oreshkin et~al.(2021)Oreshkin, Dudek, Pełka, and
  Turkina}]{oreshkin2021nbeats}
\bibinfo{author}{B.~N. Oreshkin}, \bibinfo{author}{G.~Dudek},
  \bibinfo{author}{P.~Pełka}, \bibinfo{author}{E.~Turkina},
\newblock \bibinfo{title}{{N-BEATS} neural network for mid-term electricity
  load forecasting},
\newblock \bibinfo{journal}{Applied Energy} \bibinfo{volume}{293}
  (\bibinfo{year}{2021}) \bibinfo{pages}{116918}.
  \DOIprefix\doi{10.1016/j.apenergy.2021.116918}.
%Type = Inproceedings
\bibitem[{Perez et~al.(2018)Perez, Strub, de~Vries, Dumoulin, and
  Courville}]{perez2018FiLM}
\bibinfo{author}{E.~Perez}, \bibinfo{author}{F.~Strub},
  \bibinfo{author}{H.~de~Vries}, \bibinfo{author}{V.~Dumoulin},
  \bibinfo{author}{A.~C. Courville},
\newblock \bibinfo{title}{Film: Visual reasoning with a general conditioning
  layer},
\newblock in: \bibinfo{booktitle}{AAAI}, \bibinfo{year}{2018}.
  \DOIprefix\doi{10.1609/aaai.v32i1.11671}.
%Type = Article
\bibitem[{Popławski et~al.(2015)Popławski, Dudek, and Łyp}]{Pop15}
\bibinfo{author}{T.~Popławski}, \bibinfo{author}{G.~Dudek},
  \bibinfo{author}{J.~Łyp},
\newblock \bibinfo{title}{Forecasting methods for balancing energy market in
  {P}oland},
\newblock \bibinfo{journal}{International Journal of Electrical Power and
  Energy Systems} \bibinfo{volume}{65} (\bibinfo{year}{2015})
  \bibinfo{pages}{94--101}. \DOIprefix\doi{10.1016/j.ijepes.2014.09.029}.
%Type = Incollection
\bibitem[{Paszke et~al.(2019)Paszke, Gross, Massa, Lerer, Bradbury, Chanan,
  Killeen, Lin, Gimelshein, Antiga, Desmaison, Kopf, Yang, DeVito, Raison,
  Tejani, Chilamkurthy, Steiner, Fang, Bai, and Chintala}]{paszke2019pytorch}
\bibinfo{author}{A.~Paszke}, \bibinfo{author}{S.~Gross},
  \bibinfo{author}{F.~Massa}, \bibinfo{author}{A.~Lerer},
  \bibinfo{author}{J.~Bradbury}, \bibinfo{author}{G.~Chanan},
  \bibinfo{author}{T.~Killeen}, \bibinfo{author}{Z.~Lin},
  \bibinfo{author}{N.~Gimelshein}, \bibinfo{author}{L.~Antiga},
  \bibinfo{author}{A.~Desmaison}, \bibinfo{author}{A.~Kopf},
  \bibinfo{author}{E.~Yang}, \bibinfo{author}{Z.~DeVito},
  \bibinfo{author}{M.~Raison}, \bibinfo{author}{A.~Tejani},
  \bibinfo{author}{S.~Chilamkurthy}, \bibinfo{author}{B.~Steiner},
  \bibinfo{author}{L.~Fang}, \bibinfo{author}{J.~Bai},
  \bibinfo{author}{S.~Chintala},
\newblock \bibinfo{title}{Pytorch: An imperative style, high-performance deep
  learning library},
\newblock in: \bibinfo{booktitle}{NeurIPS}, \bibinfo{year}{2019}, pp.
  \bibinfo{pages}{8024--8035}.
%Type = Misc
\bibitem[{Kingma and Ba(2015)}]{kingma2015adam}
\bibinfo{author}{D.~P. Kingma}, \bibinfo{author}{J.~Ba}, \bibinfo{title}{Adam:
  A method for stochastic optimization}, \bibinfo{year}{2015}.
  \href{http://arxiv.org/abs/1412.6980}{{\tt arXiv:1412.6980}}.
%Type = Misc
\bibitem[{Falcon and {The PyTorch Lightning team}(2019)}]{falcon2019pytorch}
\bibinfo{author}{W.~Falcon}, \bibinfo{author}{{The PyTorch Lightning team}},
  \bibinfo{title}{{PyTorch Lightning}}, \bibinfo{year}{2019}. \URLprefix
  \url{https://github.com/Lightning-AI/lightning}.
  \DOIprefix\doi{10.5281/zenodo.3828935}.
%Type = Article
\bibitem[{Dudek(2015)}]{Dud15}
\bibinfo{author}{G.~Dudek},
\newblock \bibinfo{title}{Pattern similarity-based methods for short-term load
  forecasting -- {P}art 2: {M}odels},
\newblock \bibinfo{journal}{Applied Soft Computing} \bibinfo{volume}{36}
  (\bibinfo{year}{2015}) \bibinfo{pages}{422--441}.
  \DOIprefix\doi{10.1016/j.asoc.2015.07.035}.
%Type = Inproceedings
\bibitem[{Dudek(2019)}]{Dud19}
\bibinfo{author}{G.~Dudek},
\newblock \bibinfo{title}{Short-term load forecasting using {T}heta method},
\newblock in: \bibinfo{booktitle}{14th Int. Conf. on Forecasting in Power
  Engineering 2018, E3S Web Conf.}, volume~\bibinfo{volume}{84},
  \bibinfo{year}{2019}. \DOIprefix\doi{10.1051/e3sconf/20198401004}.
%Type = Article
\bibitem[{Dudek(2016)}]{Dud16}
\bibinfo{author}{G.~Dudek},
\newblock \bibinfo{title}{Pattern-based local linear regression models for
  short-term load forecasting},
\newblock \bibinfo{journal}{Electric Power Systems Research}
  \bibinfo{volume}{130} (\bibinfo{year}{2016}) \bibinfo{pages}{139--147}.
  \DOIprefix\doi{10.1016/j.epsr.2015.09.001}.
%Type = Article
\bibitem[{Salinas et~al.(2020)Salinas, Flunkert, Gasthaus, and
  Januschowski}]{Sal20}
\bibinfo{author}{D.~Salinas}, \bibinfo{author}{V.~Flunkert},
  \bibinfo{author}{J.~Gasthaus}, \bibinfo{author}{T.~Januschowski},
\newblock \bibinfo{title}{Deepar: Probabilistic forecasting with autoregressive
  recurrent networks},
\newblock \bibinfo{journal}{International Journal of Forecasting}
  \bibinfo{volume}{36} (\bibinfo{year}{2020}) \bibinfo{pages}{1181–1191}.
  \DOIprefix\doi{10.1016/j.ijforecast.2019.07.001}.
%Type = Inproceedings
\bibitem[{Vaswani et~al.(2017)Vaswani, Shazeer, Parmar, Uszkoreit, Jones,
  Gomez, and et~al.}]{Vas17}
\bibinfo{author}{A.~Vaswani}, \bibinfo{author}{N.~Shazeer},
  \bibinfo{author}{N.~Parmar}, \bibinfo{author}{J.~Uszkoreit},
  \bibinfo{author}{L.~Jones}, \bibinfo{author}{A.~N. Gomez},
  \bibinfo{author}{et~al.},
\newblock \bibinfo{title}{Attention is all you need},
\newblock in: \bibinfo{booktitle}{Proceedings of 31st Conference on Neural
  Information Processing Systems (NIPS 2017)}, \bibinfo{year}{2017}, pp.
  \bibinfo{pages}{5998--6008}.
%Type = Misc
\bibitem[{van~den Oord et~al.(2016)van~den Oord, Dieleman, Zen, Simonyan,
  Vinyals, Graves, Kalchbrenner, Senior, and Kavukcuoglu}]{Oor16}
\bibinfo{author}{A.~van~den Oord}, \bibinfo{author}{S.~Dieleman},
  \bibinfo{author}{H.~Zen}, \bibinfo{author}{K.~Simonyan},
  \bibinfo{author}{O.~Vinyals}, \bibinfo{author}{A.~Graves},
  \bibinfo{author}{N.~Kalchbrenner}, \bibinfo{author}{A.~Senior},
  \bibinfo{author}{K.~Kavukcuoglu}, \bibinfo{title}{Wavenet: A generative model
  for raw audio}, \bibinfo{year}{2016}.
  \href{http://arxiv.org/abs/1609.03499}{{\tt arXiv:1609.03499}}.
%Type = Article
\bibitem[{Lim et~al.(2021)Lim, Arik, Loeff, and Pfister}]{Lim21}
\bibinfo{author}{B.~Lim}, \bibinfo{author}{S.~Arik},
  \bibinfo{author}{N.~Loeff}, \bibinfo{author}{T.~Pfister},
\newblock \bibinfo{title}{Temporal fusion transformers for interpretable
  multi-horizon time series forecasting},
\newblock \bibinfo{journal}{International Journal of Forecasting}
  \bibinfo{volume}{37} (\bibinfo{year}{2021}) \bibinfo{pages}{1748–1764}.
  \DOIprefix\doi{10.1016/j.ijforecast.2021.03.012}.
%Type = Book
\bibitem[{Hyndman and Athanasopoulos(2021)}]{Hyn21}
\bibinfo{author}{R.~Hyndman}, \bibinfo{author}{G.~Athanasopoulos},
  \bibinfo{title}{Forecasting: principles and practice},
  \bibinfo{publisher}{3rd edition, OTexts: Melbourne, Australia},
  \bibinfo{year}{2021}. \bibinfo{note}{Accessed on 28 December 2023}.
%Type = Article
\bibitem[{Fiorucci et~al.(2016)Fiorucci, Pellegrini, Louzada, Petropoulos, and
  Koehler}]{Fio16}
\bibinfo{author}{J.~A. Fiorucci}, \bibinfo{author}{T.~R. Pellegrini},
  \bibinfo{author}{F.~Louzada}, \bibinfo{author}{F.~Petropoulos},
  \bibinfo{author}{A.~B. Koehler},
\newblock \bibinfo{title}{Models for optimising the theta method and their
  relationship to state space models},
\newblock \bibinfo{journal}{International Journal of Forecasting}
  \bibinfo{volume}{32} (\bibinfo{year}{2016}) \bibinfo{pages}{1151--1161}.
  \DOIprefix\doi{10.1016/j.ijforecast.2016.02.005}.
%Type = Article
\bibitem[{Diebold and Mariano(1995)}]{Die95}
\bibinfo{author}{F.~X. Diebold}, \bibinfo{author}{R.~S. Mariano},
\newblock \bibinfo{title}{Comparing predictive accuracy},
\newblock \bibinfo{journal}{Journal of Business \& Economic Statistics}
  \bibinfo{volume}{13} (\bibinfo{year}{1995}) \bibinfo{pages}{253--263}.
  \DOIprefix\doi{10.1080/07350015.1995.10524599}.
%Type = Article
\bibitem[{Makridakis et~al.(2022)Makridakis, Spiliotis, Assimakopoulos, Chen,
  Gaba, Tsetlin, and Winkler}]{Mar22}
\bibinfo{author}{S.~Makridakis}, \bibinfo{author}{E.~Spiliotis},
  \bibinfo{author}{V.~Assimakopoulos}, \bibinfo{author}{Z.~Chen},
  \bibinfo{author}{A.~Gaba}, \bibinfo{author}{I.~Tsetlin},
  \bibinfo{author}{R.~L. Winkler},
\newblock \bibinfo{title}{The m5 uncertainty competition: Results, findings and
  conclusions},
\newblock \bibinfo{journal}{International Journal of Forecasting}
  \bibinfo{volume}{38} (\bibinfo{year}{2022}) \bibinfo{pages}{1365--1385}.
  \URLprefix
  \url{https://www.sciencedirect.com/science/article/pii/S0169207021001722}.
  \DOIprefix\doi{10.1016/j.ijforecast.2021.10.009}, \bibinfo{note}{special
  Issue: M5 competition}.
%Type = Article
\bibitem[{Gneiting and Raftery(2007)}]{Gne07}
\bibinfo{author}{T.~Gneiting}, \bibinfo{author}{A.~E. Raftery},
\newblock \bibinfo{title}{Strictly proper scoring rules, prediction, and
  estimation},
\newblock \bibinfo{journal}{Journal of the American Statistical Association}
  \bibinfo{volume}{102} (\bibinfo{year}{2007}) \bibinfo{pages}{359--378}.
  \DOIprefix\doi{10.1198/016214506000001437}.
%Type = Misc
\bibitem[{Tang et~al.(2022)Tang, Shen, Lin, and Huang}]{tang2022}
\bibinfo{author}{W.~Tang}, \bibinfo{author}{G.~Shen}, \bibinfo{author}{Y.~Lin},
  \bibinfo{author}{J.~Huang}, \bibinfo{title}{Nonparametric quantile
  regression: Non-crossing constraints and conformal prediction},
  \bibinfo{year}{2022}. \href{http://arxiv.org/abs/2210.10161}{{\tt
  arXiv:2210.10161}}.
%Type = Misc
\bibitem[{Park et~al.(2022)Park, Maddix, Aubet, Kan, Gasthaus, and
  Wang}]{park2022}
\bibinfo{author}{Y.~Park}, \bibinfo{author}{D.~Maddix}, \bibinfo{author}{F.-X.
  Aubet}, \bibinfo{author}{K.~Kan}, \bibinfo{author}{J.~Gasthaus},
  \bibinfo{author}{Y.~Wang}, \bibinfo{title}{Learning quantile functions
  without quantile crossing for distribution-free time series forecasting},
  \bibinfo{year}{2022}. \href{http://arxiv.org/abs/2111.06581}{{\tt
  arXiv:2111.06581}}.
%Type = Misc
\bibitem[{Fakoor et~al.(2023)Fakoor, Kim, Mueller, Smola, and
  Tibshirani}]{fakoor2023flexible}
\bibinfo{author}{R.~Fakoor}, \bibinfo{author}{T.~Kim},
  \bibinfo{author}{J.~Mueller}, \bibinfo{author}{A.~J. Smola},
  \bibinfo{author}{R.~J. Tibshirani}, \bibinfo{title}{Flexible model
  aggregation for quantile regression}, \bibinfo{year}{2023}.
  \href{http://arxiv.org/abs/2103.00083}{{\tt arXiv:2103.00083}}.
%Type = Article
\bibitem[{Chernozhukov et~al.(2010)Chernozhukov, Fernández-Val, and
  Galichon}]{chernozhukov2010}
\bibinfo{author}{V.~Chernozhukov}, \bibinfo{author}{I.~Fernández-Val},
  \bibinfo{author}{A.~Galichon},
\newblock \bibinfo{title}{Quantile and probability curves without crossing},
\newblock \bibinfo{journal}{Econometrica} \bibinfo{volume}{78}
  (\bibinfo{year}{2010}) \bibinfo{pages}{1093–1125}. \URLprefix
  \url{http://dx.doi.org/10.3982/ECTA7880}. \DOIprefix\doi{10.3982/ecta7880}.

\end{thebibliography}

% \clearpage
% \appendix

% \section{Implementation} \label{sec:implementation}

% \clearpage
% \section{Hyperparameters of the baseline models} \label{sec:baseline_models_hyperparameters}

\end{document}